%% file: main.tex
\documentclass[10pt,twocolumn,letterpaper]{article}

\usepackage[pagenumbers]{cvpr} % To force page numbers, e.g. for an arXiv version

\usepackage{graphicx}
\usepackage{amsmath}
\usepackage{amssymb}
\usepackage{booktabs}
\usepackage[sort&compress, square, numbers]{natbib}
\usepackage{subcaption}
\usepackage{enumitem}
\usepackage{xcolor}
\usepackage{balance}
\usepackage{appendix}

\usepackage[pagebackref,breaklinks,colorlinks]{hyperref}

\usepackage{mathtools}
\usepackage{bm}
\usepackage{defs-daniel}

\usepackage[capitalize]{cleveref}
\crefname{section}{Sec.}{Secs.}
\Crefname{section}{Section}{Sections}
\Crefname{table}{Table}{Tables}
\crefname{table}{Tab.}{Tabs.}

\def\cvprPaperID{4001} % *** Enter the CVPR Paper ID here
\def\confName{CVPR}
\def\confYear{2023}

\begin{document}

\title{WIRE: Wavelet Implicit Neural Representations}

\author{Vishwanath Saragadam, Daniel LeJeune, Jasper Tan, Guha Balakrishnan,\\
Ashok Veeraraghavan, Richard~G.\ Baraniuk\\
Rice University\\
{\tt\small \url{https://vishwa91.github.io/wire}}
}
\maketitle

%%%%%%%%% ABSTRACT
\begin{abstract}
   \input{abstract.tex}
\end{abstract}

%%%%%%%%% BODY TEXT
\section{Introduction}\label{sec:intro}
\input{intro}

\section{Prior Work}\label{sec:prior}
\input{prior}

\section{Wavelet Implicit Representations}\label{sec:wire}
\input{wire}

\section{Experiments}\label{sec:exp}
\input{experiments}

\section{Conclusions}\label{sec:discussions}
\input{conclusions}

\section{Acknowledgements}\label{sec:ack}
\input{ack}

\begin{appendices}
    \section{Appendix 1: Experimental Details}\label{sec:appendix1}
    \input{appendix1.tex}

\end{appendices}

%%%%%%%%% REFERENCES
\balance
{\small
\bibliographystyle{ieee_fullname}
\bibliography{refs}
}

\end{document}

%% file: abstract.tex
Implicit neural representations (INRs) have recently advanced numerous vision-related areas.
INR performance depends strongly on the choice of the nonlinear activation function employed in its multilayer perceptron (MLP) network. A wide range of nonlinearities have been explored, but, unfortunately, current INRs designed to have high accuracy also suffer from poor robustness (to signal noise, parameter variation, etc.).
Inspired by harmonic analysis, we develop a new, highly accurate and robust INR that does not exhibit this tradeoff. 
{\bf Wavelet Implicit neural REpresentation (WIRE)} uses a continuous {\bf complex Gabor wavelet} activation function that is well-known to be optimally concentrated in space-frequency and to have excellent biases for representing images.
A wide range of experiments (image denoising, image inpainting, super-resolution, computed tomography reconstruction, image overfitting, and novel view synthesis with neural radiance fields) demonstrate that WIRE defines the new state of the art in INR accuracy, training time, and robustness.

%% file: intro.tex
%{\bf INRs are bright and shiny.}
Implicit neural representations (INRs), which learn a continuous function over a set of data points, have emerged as a promising general-purpose signal processing framework.
An INR consists of a multilayer perceptron (MLP) with alternating linear layers and element-wise nonlinear activation functions.
Thanks to the MLP, INRs do not share the locality biases that can limit the performance of convolutional neural networks (CNNs).
Consequently, INRs have advanced the state of the art in numerous vision-related areas, including computer graphics~\cite{mildenhall2020nerf,mueller2022instant,kuznetsov2021neumip}, image processing~\cite{chen2021learning}, inverse problems~\cite{sun2021coil}, and signal representations~\cite{sitzmann2020implicit}.

\begin{figure}[!tt]
    \centering
    \includegraphics[width=\columnwidth]{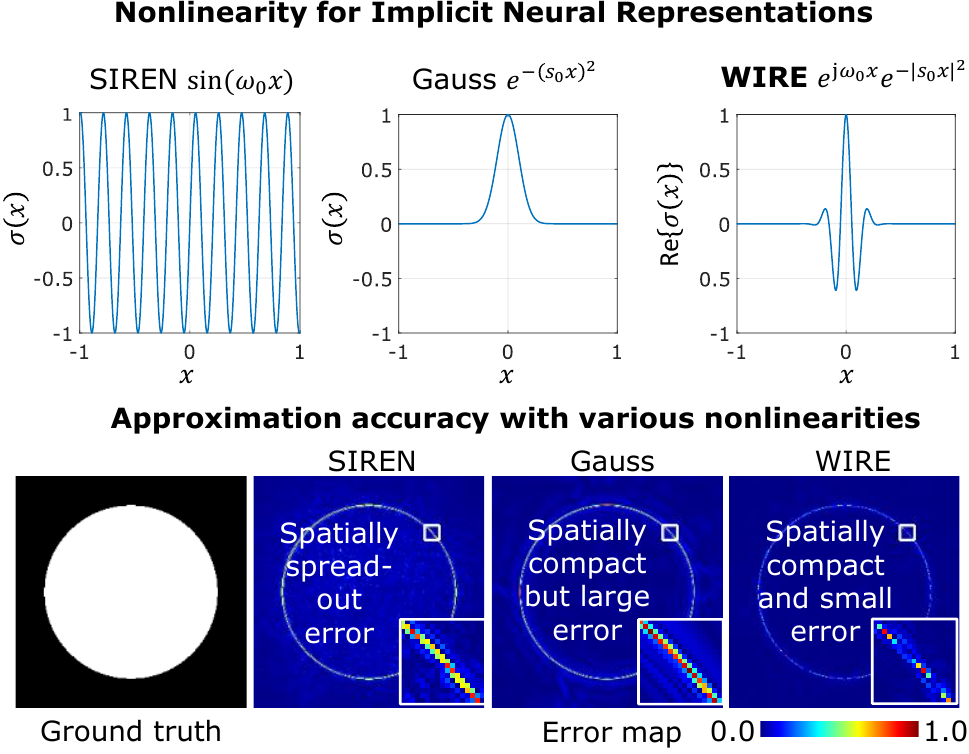}
    \caption{\textbf{Wavelet implicit neural representation (WIRE).} We propose a new nonlinearity for implicit neural representations (INRs) based on the continuous complex Gabor wavelet that has high representation capacity for visual signals. The top row visualizes two commonly used nonlinearities: SIREN with sinusoidal nonlinearity and Gaussian nonlinearity, and WIRE that uses a continuous complex Gabor wavelet. WIRE benefits from the frequency compactness of sine, and spatial compactness of a Gaussian nonlinearity. The bottom row shows error maps for approximating an image with strong edges. SIREN results in global ringing artifacts while Gaussian nonlinearity leads to compact but large error at edges. WIRE produces results with the smallest and most spatially compact error. This enables WIRE to learn representations rapidly and accurately, while being robust to noise and undersampling of data.}
    \label{fig:teaser}
\end{figure}

% current INRs are still just meh
Currently, INRs still face a number of obstacles that limit their use. First, for some applications, especially those with high-dimensional data such as 3D volumes, fitting an INR to high accuracy can still take too long (tens of seconds) for real time applications. Second, INRs are not robust to signal noise or insufficient measurements. Indeed, most works on INRs in the literature assume virtually no signal noise and large amounts of data. We find in our own experiments that current INR methods are ineffective for tasks such as denoising or super-resolution. Finally, INRs still have room for improvement in representational accuracy, especially for fine details.

%{\bf Menagerie of activations and the key tradeoff.}
%INR performance depends strongly on the choice of activation function.
%A wide range of nonlinearities have been explored, including piecewise-linear ReLU functions \cite{X}, Gaussian bumps \cite{X}, and sinusoids \cite{X}. 
%Unfortunately, empirical explorations \jaspo{ours or in literature?} have pointed towards what seems to be a difficult tradeoff in INR design: INRs designed to have high accuracy (e.g., sine activation) seem to suffer from poor robustness (to signal noise, variation in the sine parameters, etc.).

%{\bf Inspiration from harmonic analysis.}
%In this paper, we develop a new, highly accurate, and robust INR that does not suffer from this tradeoff.
In this paper, we develop a new, faster, more accurate, and robust INR that addresses these issues and takes INR performance to the next level.
To achieve this, we take inspiration from harmonic analysis and reconsider the nonlinear activation function used in the MLP.
% Our design inspiration is harmonic analysis.
Recent work has shown that an INR can be interpreted as a structured signal representation dictionary~\cite{yuce2022structured}, where the activation nonlinearity dictates the atoms of the dictionary.
For example, the sine activation creates a pseudo-Fourier transform representation of the signal that is maximally concentrated in the frequency domain~\cite{yuce2022structured}.

%{\bf But, if sines are good, then wavelets are better!}
An important conclusion one can draw from the past four decades of harmonic analysis research is that Fourier methods are suboptimal for representing the kinds of signals that feature in typical vision tasks \cite{mallat1999wavelet}.
These kinds of signals, e.g., natural images from photographs, are much more concisely and robustly represented using {\em wavelet} atoms that are optimally concentrated in space--frequency.
Sparse compositions of wavelet atoms are known to have excellent biases for representing images; cf.\ the seminal work in computer vision (e.g., Laplacian pyramid), computational neuroscience~\cite{olshausen1996emergence}, and the JPEG2000  compression standard.

%{\bf Hanging by a WIRE.}
In this paper, we introduce {\bf Wavelet Implicit neural REpresentation (WIRE)}, a new INR based on a {\em complex Gabor wavelet} activation function (see Figure~\ref{fig:teaser}).
Through a wide range of experiments, we demonstrate that WIRE defines the new state of the art in INR accuracy, training time, and robustness.
We showcase that WIRE's increased robustness is particularly useful for solving difficult vision inverse problems, including image denoising (robustness), image inpainting and super-resolution (superior interpolation), and 2D computed tomography (CT) reconstruction (solving higher-dimensional inverse problems). 
WIRE also outperforms other INRs for signal representation tasks such as overfitting images and learning point cloud occupancy volumes.
Finally, we show that WIRE enables faster, more robust novel view synthesis with neural radiance fields (NeRF)~\cite{mildenhall2020nerf} from critically few training views.

%% file: prior.tex
\paragraph{Regularization for inverse problems.} Inverse problems involve estimating a signal from a linear or nonlinear set of measurements. 
Inevitably, the measurements are degraded by noise (such as camera readout or photon noise), or the problem is ill-conditioned, necessitating regularization.
There are many forms of regularization, including ridge regression, Lasso~\cite{tibshirani1996regression}, total variation (TV)~\cite{chambolle2004algorithm}, and sparsity-based~\cite{baraniuk2010model} techniques that seek to penalize the $\ell_1$ norm the signal or some transform thereof.
In the past decade, data-driven regularization, including overcomplete dictionary-based~\cite{aharon2006k} and generative network-based~\cite{nguyen2017plug,romano2017little,rick2017one} ones, have been developed.
The classical model-based approaches are inadequate for severely ill-conditioned problems, while the data-driven ones critically depend on data.

%\richb{richb: i don't think this para on CNNs is relevant to the intro; i would move to a CNN paragraph in section 2 prior work}
\paragraph{Convolutional neural networks (CNNs).} CNNs, the most popular neural network architectures in computer vision for the past decade, have been shown to exhibit strong implicit biases that favor image-like signals.
This has been demonstrated with works like deep image prior (DIP)~\cite{ulyanov2018deep} and its variations~\cite{heckel2018deep,darestani2021untrained} that produce remarkable results on image-related linear inverse problems without any prior training data.
However, such CNN-based priors are tied to a discrete grid-like signal representation which is not applicable to problems such as novel view synthesis, or for solving ordinary and partial differential equations, and not scalable for very high dimensional signals such as 3D tomographic volumes, gigapixel images, or large point clouds.

\paragraph{Deep image prior.} Neural networks, and particularly CNNs, exhibit implicit biases due to their specific architectures (such as a UNet~\cite{ronneberger2015u}), implying that even untrained neural networks can be used for regularization.
This was leveraged to build a deep image prior (DIP)~\cite{ulyanov2018deep} that produces outputs that tend to look like images.
The key idea is to recast regularization as optimizing for the weights of the network for each instance of the problem.
The performance of DIPs is considerably superior to classical regularization approaches.
However, DIPs exhibit good performance only when over-parameterized  and are tied to a grid-like discretized representation of the signal, implying DIPs do not scale to high dimensional signals such as point clouds with a large number of points.
The issue of computational cost has been addressed to a certain extent by the deep decoder~\cite{heckel2018deep} and the DeepTensor~\cite{saragadam2022deeptensor}, but they still need the signal to be defined as a regular data grid such as a 2D matrix or 3D tensor.

\paragraph{Implicit representations.} INRs are continuous learned function approximators based on multilayer perceptrons (MLPs).
The continuous nature of INRs is particularly appealing when dealing with irregularly sampled signals such as a point clouds.
Since its first widespread usage in novel view synthesis in graphics~\cite{mildenhall2020nerf}, INRs have pervaded nearly all fields of vision and signal processing including rendering~\cite{kuznetsov2021neumip}, computational imaging~\cite{attal2021torf,chien1986volume}, medical imaging~\cite{wang2022neural}, and virtual reality~\cite{deng2022fov}.

The popular choice of the ReLU nonlinearity in standard neural networks has been empirically shown to result in poor approximation accuracy in INRs.
This has been remedied by several modifications to the MLP including the so-called positional encoding~\cite{tancik2020fourfeat,mueller2022instant}, as well as various choices of nonlinearity such as the sinusoidal function~\cite{sitzmann2020implicit} and the Gaussian function~\cite{ramasinghe2021beyond}.
%
%, and a multiplicative frequency network (MFN)~\cite{fathony2020multiplicative}.
%\daniel{somehow I think we must acknowledge that MFN uses Gabor}
%
A closely related work is the Gabor wavelet-based multiplicative filter networks (MFN), where the output after each layer is multiplied by a Gabor filter.
The output then results in a combination of exponentially many Gabor wavelets, thereby resulting in large capacity.
Numerous architectural changes have also been proposed that leverage multiscale properties of visual signals to accelerate the INR training procedure including adaptive block decomposition~\cite{martel2021acorn}, kilo-NeRF~\cite{reiser2021kilonerf}, and predicting the Laplacian pyramid of the signal~\cite{saragadam2022miner}.

INRs can now train on signals nearly instantly~\cite{mueller2022instant} thanks to these numerous advances. 
However, the high capacity of such INRs precludes robustness --- implying that the signal representation is brittle, resulting in overfitting to both noise and signal equally. %\jaspo{Current INRs overfitting to noise is an argument used throughout the paper, but I feel it is lacking in evidence in this paper. Is this well-known in literature? Can we justify this argument (perhaps provide references that show this)?}
In this paper, we propose the complex Gabor wavelet as a nonlinearity, which is uniquely well-suited to induce robustness in INRs.

\paragraph{Wavelet transform and the Gabor wavelet.}
%\richb{Gabor wavelet?}
%\vishwa{Need to think of what is the best way to write this.}
%
The Fourier transform decomposes the signal as a sum of sinusoids with infinite space support, implying that there is no notion of spatial compactness.
The wavelet transform remedies this by decomposing the signal as a linear combination of translated and scaled versions of a short oscillating pulse called a wavelet.
Wavelets typically result in faster approximation rates for signals and images than Fourier transform~\cite{devore_1998}, and hence they are often used for image compression~\cite{devore1992image,shapiro1993embedded} and as a robust prior for inverse problems of images~\cite{he2009exploiting} and videos~\cite{wakin2006compressive}.
In this paper, we show that wavelets are a universally superior choice for the nonlinearity in INRs due to their compact support in space and frequency and therefore faster approximation rates.

%% file: wire.tex
%\begin{figure*}[!tt]
%    \centering
%    \includegraphics[width=\textwidth]{figures/nonlin.pdf}
%    \caption{\textbf{Comparison of nonlinearities.} We propose a new nonlinearity based on the continuous Gabor wavelet that is compact like the Gaussian nonlinearity, has oscillations like SIREN that enables it to approximate high frequencies, and is infinitely differentiable.}
%    \label{fig:nonlin}
%\end{figure*}

\subsection{INR details}
Consider an INR function $F_\theta: \mathbb{R}^{D_i} \mapsto \mathbb{R}^{D_o}$ mapping $D_i$ input dimensions to $D_o$ output dimensions, where $\theta$ represents the MLP's tunable parameters. The goal is to construct $F_\theta$ such that it approximates a function $g(\vx)$ of interest, i.e., $g(\mathbf{x}) \approx F_\theta(\mathbf{x})$. For example, $g(\vx)$ may simply be a ground truth image, represented as a function mapping coordinates to pixel values.
Modeling $F_\theta (\cdot) $ as an $M$-layer MLP, the output at each layer is given by
\begin{align}
    \mathbf{y}_{m} = \sigma(W_{m}\mathbf{y}_{m-1} + \mathbf{b}_{m}),
\end{align}
where $\sigma$ is the nonlinearity (or nonlinear activation function);
$W_m, \mathbf{b}_m$ are weights and biases for the $m^\text{th}$ layer;
$\mathbf{y}_0 = \mathbf{x} \in \reals^{D_i}$ is the input coordinate and $\vy_{M+1} = W_{M+1}\vy_M + \vb_{M+1}$ is the final output.

The nonlinear activation $\sigma$ plays a key role in the representation capacity of the INR (see Fig.~\ref{fig:teaser}).
Two leading choices include
the periodic $\sigma(x) = \sin(\omega_0 x)$ used in 
SIREN~\cite{sitzmann2020implicit}, 
and the Gaussian nonlinearity $\sigma(x) = e^{-(s_0x)^2}$
used by Ramasinghe et.~al.~\cite{ramasinghe2021beyond}---both result in significantly higher representation accuracy than ReLU.
%uses a  or a similar spatially compact nonlinearity
%
However, their high representation capacity is also a drawback, since they can represent noise with nearly equal accuracy as an image.
Our goal is to propose a nonlinearity $\sigma$ that is well-suited for visual signals such as images, videos, and 3D volumes but poorly fits noise-like signals.
%
%\vishwa{I think we need some NTK-like argument here.}
%\daniel{Can you elaborate on what is needed? I can jump in.}
%\vishwa{I was hoping to show that SIREN and Gauss do not care about images and hence are good for all classes of signals. However, that is bad because noise and image are the same to SIREN/Gauss. NTK migh reveal this.}

\subsection{WIRE}

Armed with the insight that a Gabor wavelet achieves optimal time-frequency compactness, we propose the wavelet implicit representation (WIRE) that uses the \textbf{continuous complex Gabor wavelet $\psi$ for its activation nonlinearity}:
\begin{align}
    \sigma(x) = \psi(x; \omega_0, s_0) = e^{j\omega_0x}e^{-|s_0x|^2}, 
\end{align}
where $\omega_0$ controls the frequency of the wavelet and $s_0$ controls the spread (or width). The first layer activations have the form
\begin{align}
    \mathbf{y}_1 = \psi(W_1 \mathbf{x} + \mathbf{b}_1; \omega_0, s_0),
\end{align}
which are copies of the mother Gabor wavelet $\psi$ at scales and shifts determined by $W_1$ and $\mathbf{b}_1$. Hence the building blocks of WIRE are drawn from a dictionary of wavelet atoms.
%Since noise does not enjoy compact spatial support, we can expect WIRE to represent visual signals with much higher fidelity than noise, thereby resulting in an implicit bias.
%
We let the weights of the INR as well as the outputs be complex-valued to preserve phase relationships throughout,
and we represent real signals by simply taking the real part of the output and discarding the imaginary part.
Just as wavelets combine space and frequency compactness, WIRE enjoys the advantages of periodic nonlinearities such as SIREN due to the complex exponential term and the spatial compactness from the Gaussian window term; recall Figure~\ref{fig:teaser}. 
%\daniel{say WIRE doesn't need weird SIREN init somewhere?}
Additionally, unlike SIREN, WIRE does not require a carefully chosen set of initial weights (see Fig.~\ref{fig:imsweep})  due to the Gaussian window, which creates a spatially compact output at each layer and produces high quality results with the default neural network initialization of uniformly random weights independent of the parameters $\omega_0, s_0$.

\begin{figure}[!tt]
    \centering
    \begin{subfigure}[t]{0.47\columnwidth}
        \centering
        \includegraphics[width=\columnwidth]{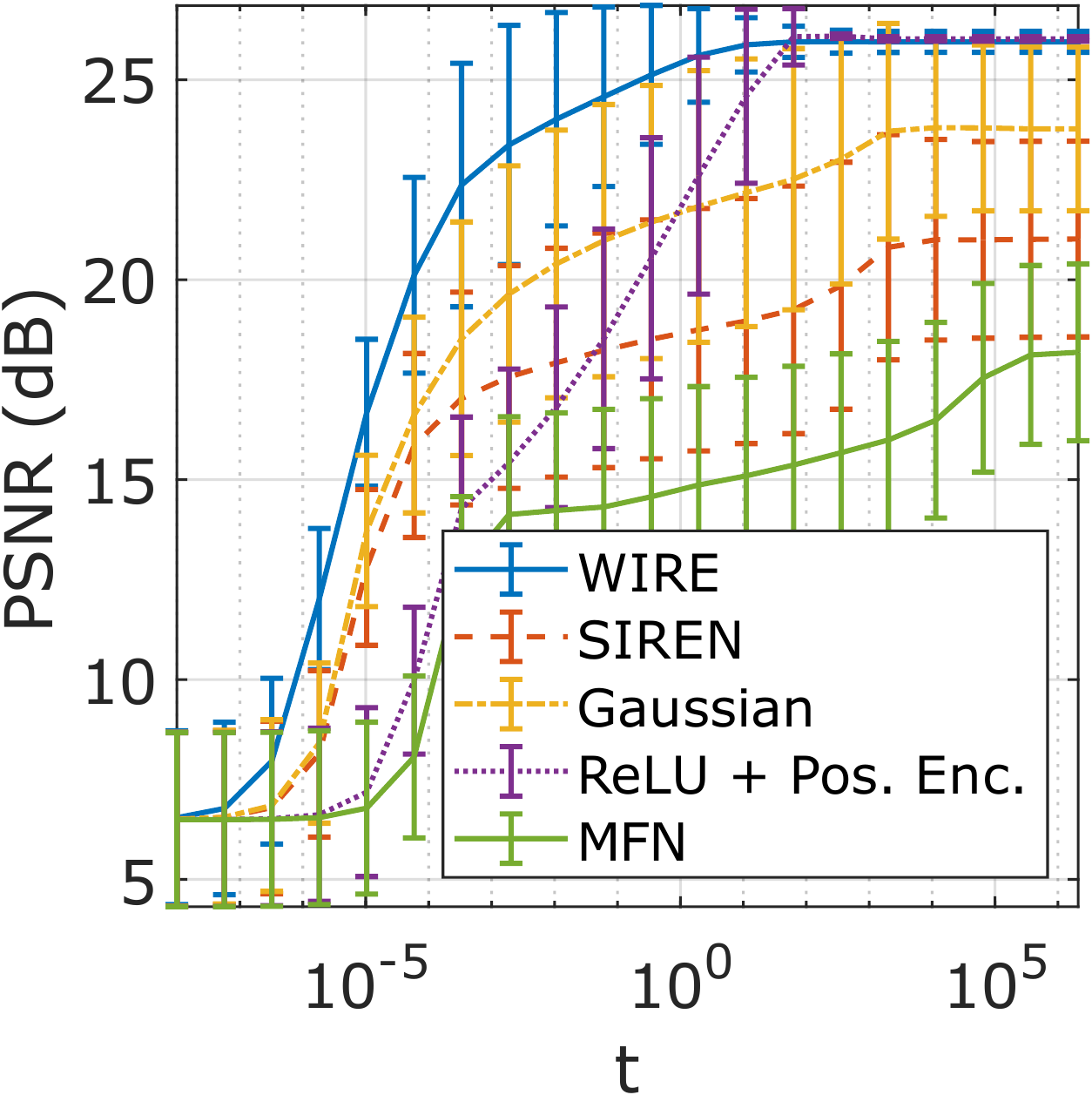}
        \caption{\centering Implicit bias with NTK gradient flow}
        \label{fig:implicit-ntk}
    \end{subfigure}
    \hspace{0.2em}
    \begin{subfigure}[t]{0.48\columnwidth}
        \centering
        \includegraphics[width=\columnwidth]{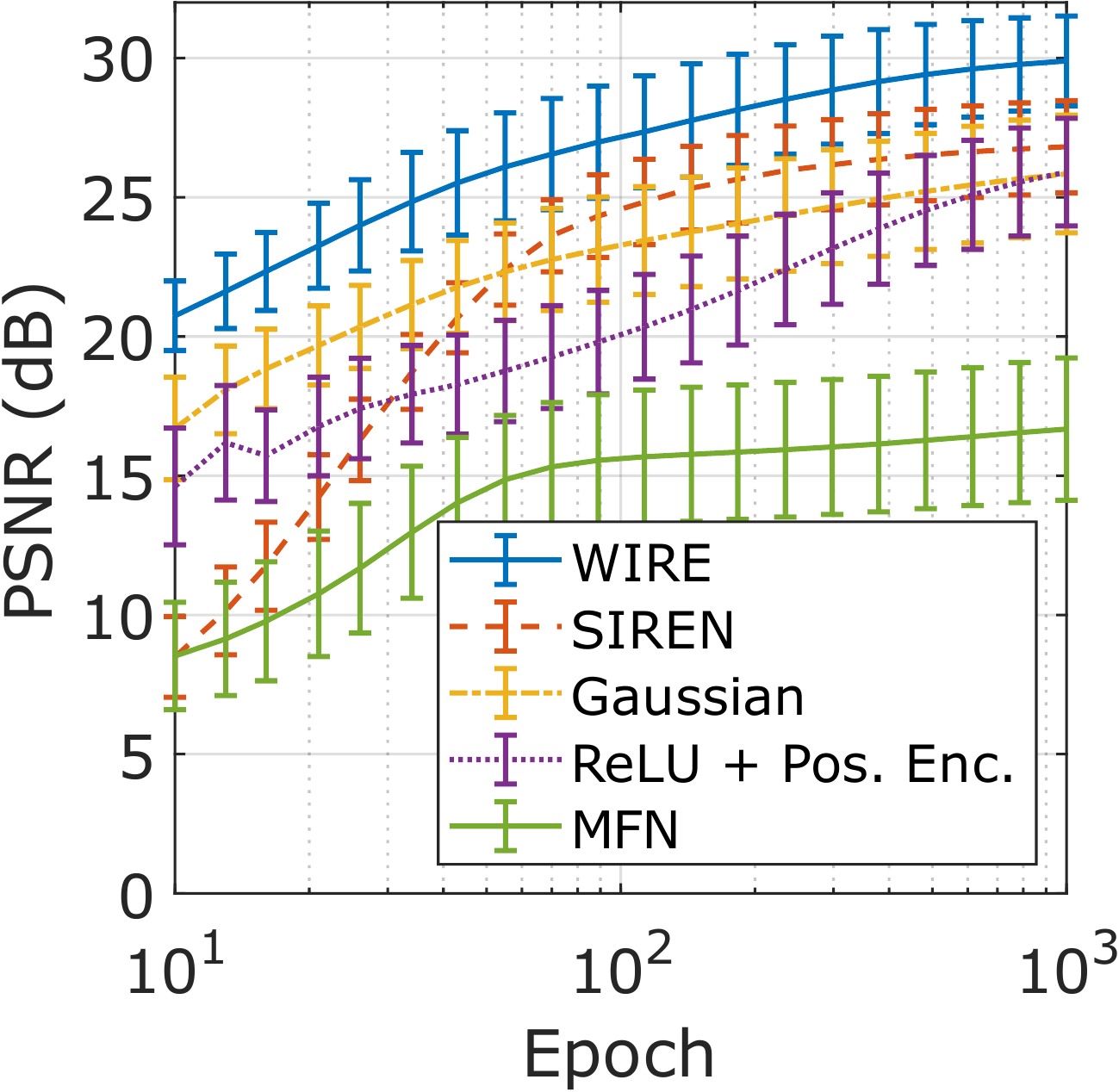}
        \caption{\centering Implicit bias in standard INRs.}
        \label{fig:implicit-empirical}
    \end{subfigure}
    \caption{\textbf{Implicit bias in denoising} (a) The empirical NTK of finite-width INRs provides an insight into the implicit bias of INRs. Early trajectories of NTK gradient flow show WIRE converging to the image faster than the noise, outperforming all other nonlinearities. Bars indicate one standard deviation over the dataset. (b) Early iterations of standard training are reflected well by the relative performances of NTK gradient flow from part (a). Furthermore, WIRE maintains its advantage against other nonlinearities throughout the remainder of training.
    %The images on top show the top 64 NTK eigenvectors on $[-1, 1]^2$ for untrained INRs with ReLU, sine, Gauss, and Gabor (WIRE) nonlinearities, while the bottom plot shows image approximation accuracy using gradient flow over the Tiny Imagenet dataset. WIRE achieves higher reconstruction accuracy early on, implying that WIRE has implicit bias that is well-suited for representing images. \daniel{re-do this caption once figure is finalized}
    }
    \label{fig:implicit}
\end{figure}

%\begin{figure}[!tt]
%    \centering
%    \includegraphics[width=\columnwidth]{figures/implicit_bias.pdf}
%    \caption{\textbf{INRS with WIRE favors visual signals.} The plots above show training accuracy for an image and uniformly random noise. WIRE shows a large difference in approximation of the two signals, thereby exhibiting implicit bias. In contrast, SIREN, Gauss, and ReLU with positional encoding show lower separation between image and noise for the first 100 epochs.}
%    \label{fig:implicit}
%\end{figure}

\subsection{Implicit bias of WIRE}
\paragraph{Neural tangent kernel perspective.} As stated, we seek an INR that fits visual signals well but fits noise poorly in comparison. 
Inspired by \cite{yuce2022structured}, who proposed to compare eigenfunctions of the empirical neural tangent kernel (NTK)~\cite{jacot2018ntk} of INRs to understand their approximation properties, we compare the fitting of noisy natural images using NTK gradient flow. 
The NTK gradient flow of INRs accurately captures the behavior of early training of neural networks, and so in tasks such as denoising where we regularize via early stopping, the early training behavior determines the implicit bias. In the lazy training regime of wide neural networks~\cite{lee2019wide}, the fit image at time $t \geq 0$ has value
\begin{align}
    F_{\theta_t}(\mathbf{x}) = [(I - e^{-t K}) g](\mathbf{x}),
\end{align}
where $I$ is the identity operator, $K$ is the NTK operator on the image's spatial domain, and $g$ is the image being fit.

In Fig.~\ref{fig:implicit-ntk}, we apply NTK gradient flow using the empirical finite-width NTK to a denoising task, fitting the original image with $\mathcal{N}(0, 0.05^2)$ i.i.d.\ pixel-wise additive noise. Due to the computational intensity of evaluating the NTK, we evaluate on $64 \times 64 \times 3$ images from Tiny ImageNet~\cite{tinyimagenet2015}.
Comparing WIRE to other INRs, we see that, as desired, WIRE prefers to learn the signal in the image early in training rather than the noise, converging orders of magnitude faster to essentially any given peak signal-to-noise-ratio (PSNR). % \daniel{in early training}.

% \daniel{I can elaborate more here where needed / if there is space}
% Since WIRE is equipped with Gabor nonlinearity, we expect its implicit bias to favor images.

% To test our hypothesis, we evaluated the empirical neural tangent kernel (NTK)~\cite{jacot2018ntk} for INRs as proposed in~\cite{yuce2022structured} for various nonlinearities.

% We first computed the eigenvectors for the empirical NTK supported on $[-1, 1]^2$ of INRs equipped with ReLU, sine, Gaussian, and Gabor nonlinearities.

% Then, using gradient flow dynamics, we computed approximation accuracy over iterations over the images in the Tiny ImageNet dataset \cite{tinyimagenet2015}.

% Figure~\ref{fig:ntk} shows the top eigenvectors for each nonlinearity, and the gradient flow dynamics.
% \daniel{we may update this to include dynamics on noise / denoising?}

% We notice that WIRE equipped with Gabor nonlinearity outperforms all other nonlinearities over images, thereby confirming our hypothesis that WIRE's implicit bias is well-suited for images.
% \daniel{converges orders of magnitude faster}

\paragraph{Empirical evaluation.} We perform a denoising task analogous to the NTK-based analysis for real INRs on the 24 $768 \times 512 \times 3$ images from the Kodak Lossless True Color Image Suite~\cite{kodak1999}, again with $\mathcal{N}(0, 0.05^2)$ additive noise, in Fig.~\ref{fig:implicit-empirical}.
We apply the same INRs as in the NTK example, but train with ordinary neural network gradient optimization instead of NTK gradient flow.
Again, WIRE drastically outperforms other INRs, converging an order of magnitude faster to the same PSNR.
%To test our hypothesis that WIRE represents images more accurately than noise, we trained INRs equipped with WIRE, SIREN, Gauss, and ReLU with positional encoding nonlinearities on an image and uniformly random noise.
%
%The two signals and the training mean-squared error (MSE) across epochs is visualized in

%
%We notice that training accuracy with WIRE has an increasing difference between noise and image while other nonlinearities have smaller speration for nearly 100 epochs. This observation implies that the final result for an addition of the two images will be noisier with other nonliniearites.

%\subsection{Understanding WIRE parameters}
%\daniel{weird title?}
%

% \subsection{Space--frequency localization} 
% \vishwa{I do not have conclusive results for this, we need to discuss.}
% WIRE inherits a key advantage of decomposing signals with wavelets, namely the space-frequency localization.
% %
% This is particularly appealing as visual signals tend to have compact support of high frequency terms in the form of edges.
% %
% \daniel{Connect to Figure 1?}

\subsection{Choosing the parameters $\omega_0, s_0$}

\begin{figure}[!tt]
    \centering
    \includegraphics[width=\columnwidth]{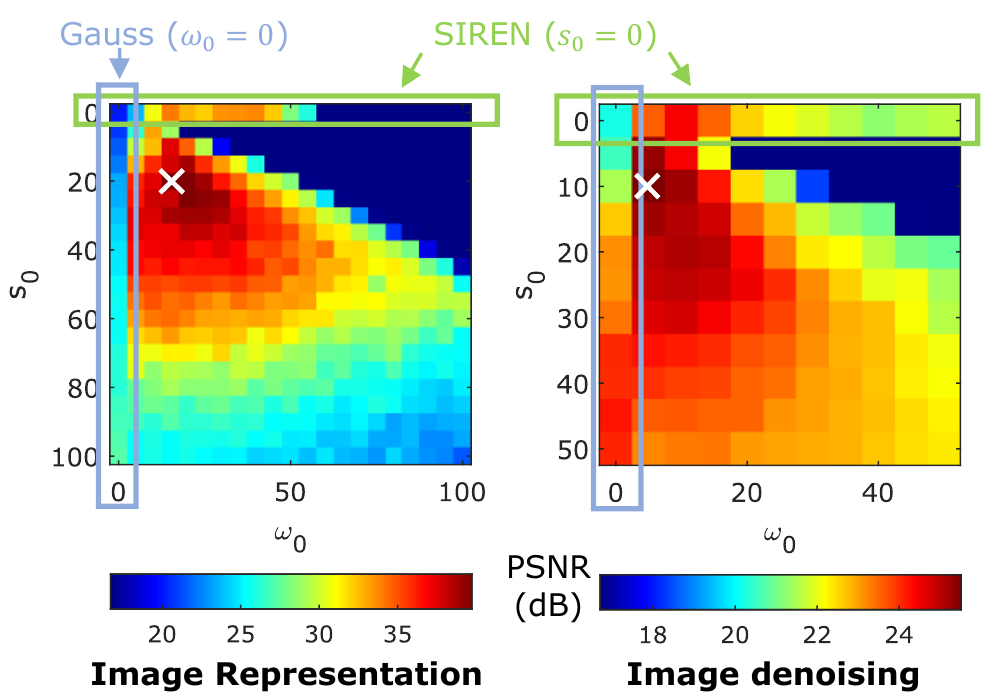}
    \caption{\textbf{WIRE is robust to the choice of parameters.} The plot above shows accuracy for image representation and denoising with various settings of $\omega_0$ and $s_0$. The boxes show special cases with $\omega=0$ corresponding to Gaussian nonlinearity, and $s_0=0$ corresponding to SIREN. WIRE achieves higher accuracy than both SIREN and Gauss on image representation as well as image denoising tasks (marked by white cross). Further, WIRE achieves super performance for a large choice of parameters $\omega_0, s_0$ implying that WIRE is not overly sensitive to the hyperparameters.}
    \label{fig:imsweep}
\end{figure}

WIRE's performance is primarily decided by the constants $\omega_0, s_0$ that control frequency of the sinusoid and width of the Gaussian, respectively.
WIRE outperforms both the SIREN and Gaussian nonlinearities across a broad range of parameters.
Figure~\ref{fig:imsweep} shows the approximation accuracy achieved by WIRE for various parameters.
We set the number of hidden layers to three, and number of hidden features to 256.
When $\omega_0 = 0$, we used a Gaussian nonlinearity. When $s_0 = 0$, we used a sinusoidal nonlinearity.
When both parameters were zero, we used a ReLU nonlinearity.
For the denoising task, we added photon noise equivalent to a maximum of 50 photons per pixel. 
We observe from Fig.~\ref{fig:imsweep} that WIRE outperforms SIREN, Gauss, and ReLU.
Moreover, the performance is superior for a large swath of values of $\omega_0, s_0$ for both image representation and denoising.
The reduced sensitivity to the exact values of $\omega_0, s_0$ implies that WIRE can be used without precise information about image or noise statistics.

\paragraph{Alternate forms of WIRE.} For problems where complex weights are infeasible, WIRE can be instantiated as the imaginary (or real) part of the complex Gabor wavelet, $\psi(x; \omega_0, s_0) = \sin(\omega_0x)e^{-(s_0x)^2}$. Note that setting $s_0 = 0$ results in the sine nonlinearity used in SIREN~\cite{sitzmann2020implicit} and $\omega_0 = 0$ results in Gaussian nonlinearity~\cite{ramasinghe2021beyond} implying WIRE inherits the favorable properties of previously proposed nonlinearities.
Another embodiment of WIRE is a Constant-$Q$ Gabor wavelet where $\omega_0s_0 = Q$, which results in constant fractional bandwidth ($\omega/\delta\omega)$. Constant-$Q$ Gabor wavelets are often used in music analysis~\cite{brown1991calculation,todisco2017constant}, wavelet transforms~\cite{mallat1999wavelet}, and the Laplace transform. Having only a single parameter makes hyperparameter tuning simpler with a fixed $Q$.

\subsection{Multidimensional localization}\label{sec:multi}

As defined above, WIRE applies the Gabor mother wavelet $\psi$ element-wise to output of the linear transformation $W_m \vy_{m-1}$.
%\richb{richb attempt to reword what's commented out below:}
Hence, the output of each unit is spatially localized only in the single direction determined by the corresponding row of $W_m$ (see Fig.~\ref{fig:wire_2d_act} top).
%
%Hence it provides spatial localization only along the dimension \richb{do you mean `direction'?} of that transformation determined by the rows of $W_m$.
%
While such highly anisotropic spatial localization is well-suited for certain kinds of data, many natural data (e.g., photographic images) are best represented using a combination of atoms with isotropic and anisotropic spatial localization (e.g., wavelets and curvelets \cite{candes2004new} or wavelets and wedgelets \cite{wakin2006wavelet}).
To achieve spatial localization along multiple directions, we augment the Gabor mother wavelet with $D_m-1$ additional Gaussian windows:
\begin{equation}
    \begin{aligned}
    \vy_m ={} &\psi(W_m^{(1)} \vy_{m-1} + \vb_m^{(1)}; w_0, s_0) \\
    &~~~\cdot e^{-\sum_{k=2}^{D_m} |s_0 (W_m^{(k)} \vy_{m-1} + \vb_m^{(k)})|^2}.
    \end{aligned}
\end{equation}
In two-dimensional settings, such as with natural image data, the resulting first-layer activations will resemble a mixture of Gabor wavelets and curvelets (see Fig.~\ref{fig:wire_2d_act} bottom).
As we will see below in Section~\ref{sec:multi-perf}, the more diverse spatial localization of the resulting 2D WIRE representation significantly benefits its performance (see Fig.~\ref{fig:wire_2d_img} and Fig.\ref{fig:wire_2d_comp}).

%% file: experiments.tex
%Spoiler alert!  
WIRE learns representations for all signal classes faster than state-of-the-art techniques.
In addition, WIRE is well-suited to solve a large class of inverse problems where the number of measurements is far fewer than the dimensionality of the signal, or when the measurements are corrupted by noise.
For all the experiments below, we implemented the optimization procedure in PyTorch~\cite{NEURIPS2019_9015} and used the Adam optimizer~\cite{kingmaB14}.
Code was executed on a system unit equipped with 64GB RAM, and an Nvidia RTX 2080 Ti graphical processing unit (GPU) with 8GB memory.
Unless specified, we used an $\ell_2$ loss function between the measurements and the outputs of INR.
No other regularization was used.
We used a learning rate scheduler which decayed the initial learning rate by $0.1$ at the end of training epochs.

\begin{figure}[!tt]
    \centering
    \begin{subfigure}[t]{0.48\columnwidth}
        \centering
        \includegraphics[width=\textwidth]{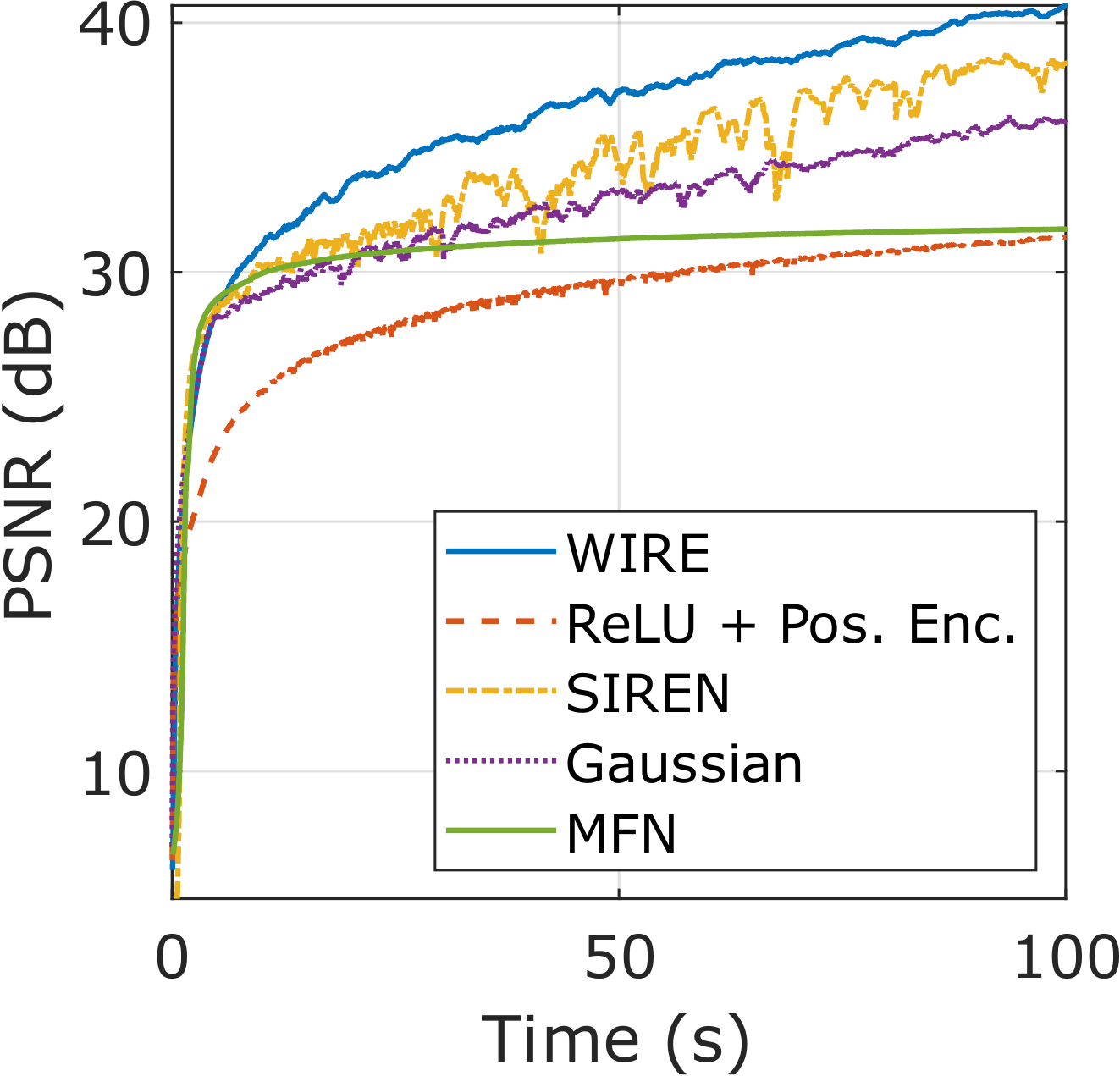}
        \caption{Image representation}
    \end{subfigure}
    \begin{subfigure}[t]{0.49\columnwidth}
        \centering
        \includegraphics[width=\textwidth]{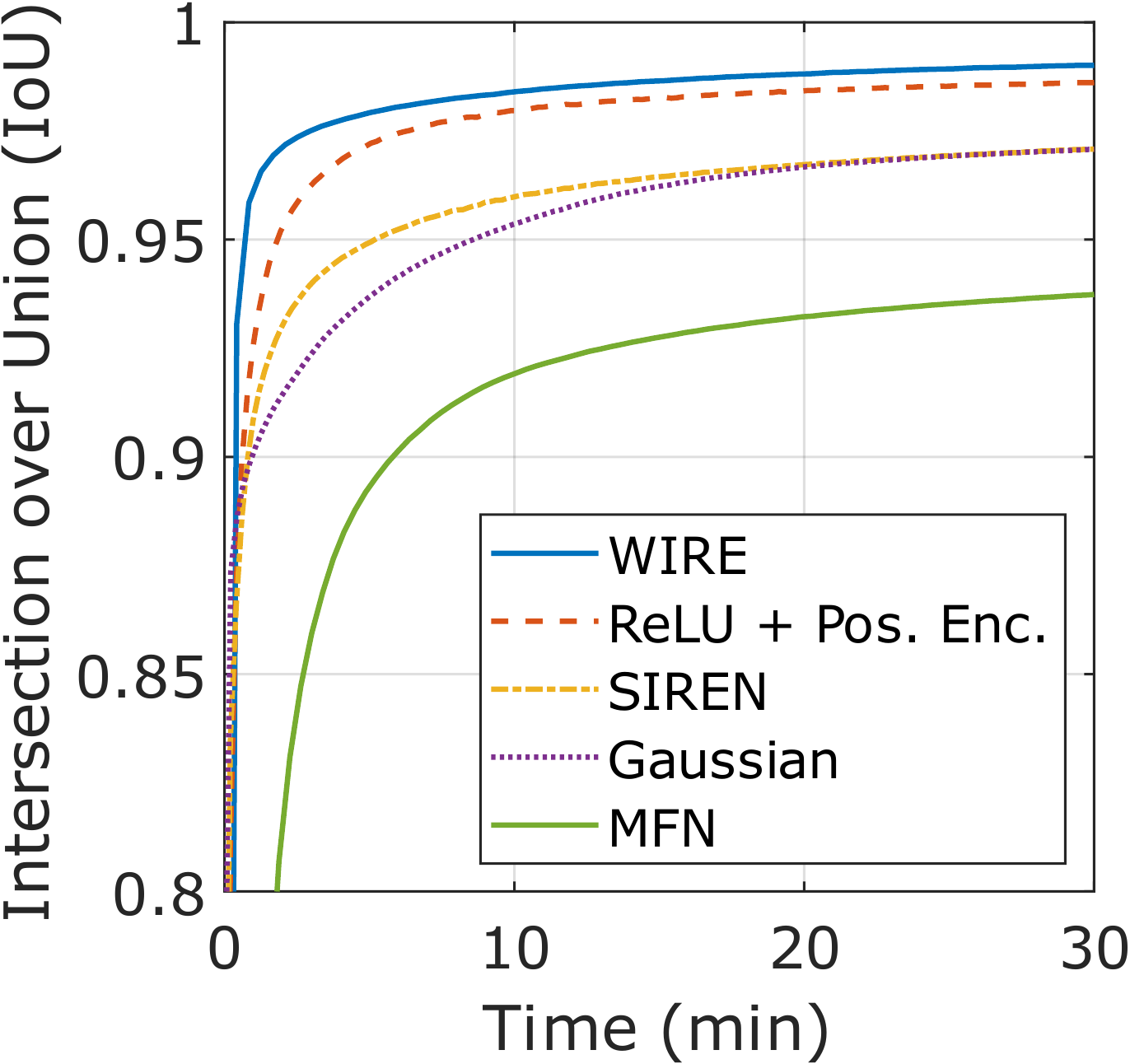}
        \caption{Volume representation}
    \end{subfigure}
    \caption{\textbf{WIRE learns faster.} The two plots above show representation accuracy for an image (top row in Fig.~\ref{fig:rep}) and an occupancy volume (bottom row in Fig.~\ref{fig:rep}) over time. Owing to the high approximation capacity of Gabor wavelets for visual signals, WIRE achieves high accuracy at a faster rate, making it an appropriate choice for representing visual signals.}
    \label{fig:epochs}
\end{figure}

\begin{figure*}[!tt]
    \centering
    \includegraphics[width=0.95\textwidth]{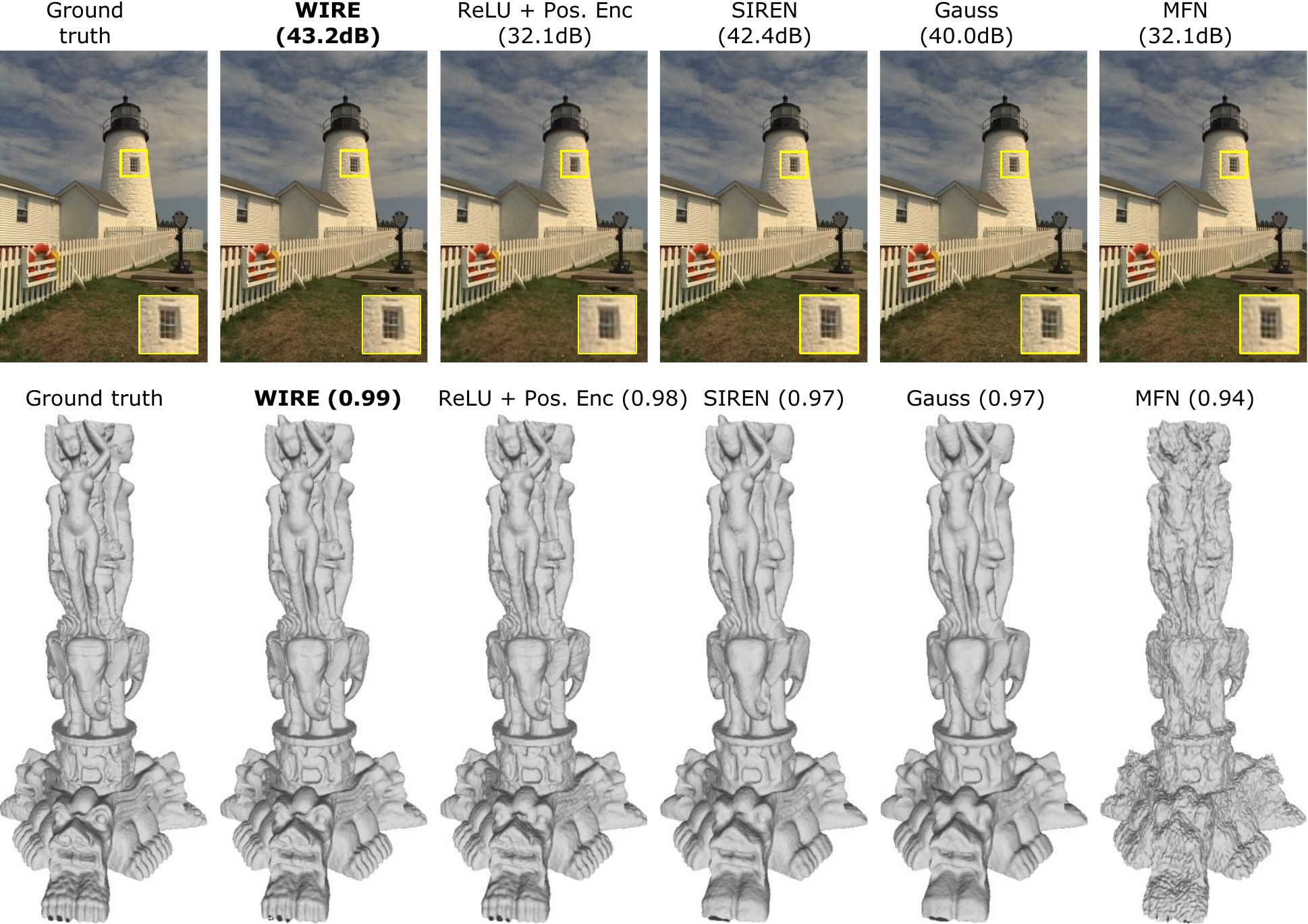}
    \caption{\textbf{WIRE has high representation capacity.} The results above show image representation in the first row and meshes generated with occupancy volumes in the second row with various nonlinearities. WIRE achieves highest representation accuracy for both data, underlining its advantages as a signal model.}
    \label{fig:rep}
\end{figure*}

\subsection{Signal representation}
A common feature enabled by INRs is representation of signals.
We evaluate two tasks for this experiment: representing images and representing occupancy volumes~\cite{mescheder2019occupancy}.
In both cases, we used an MLP with three hidden layers with a width of 300 features for all nonlinearities.
For WIRE, we reduced the number of parameters by half to account for the doubling due to real and imaginary parts.
We did so by reducing the number of hidden features by $\sqrt{2}$.
The parameters for each nonlinearity and the learning rate were chosen to obtain fastest approximation rate.
Specifically, we chose $\omega_0=20, s_0=10$ for WIRE, $\omega_0=40$ for SIREN, and $s_0=30$ for Gaussian.
We also compare against multiplicative frequency networks (MFN)~\cite{fathony2020multiplicative}.
For the occupancy volume, we sampled over a $512\times512\times512$ grid with each voxel within the volume assigned a 1, and voxels outside the volume assigned a 0.
We evaluated the PSNR and structural similarity (SSIM)~\cite{wang2004image} for images and intersection over union (IOU) for the occupancy volumes.

Figure~\ref{fig:epochs} shows the approximation accuracy as a function of time for an image (Kodak dataset) and an occupancy volume (Thai statue).
WIRE not only achieves the highest accuracy, but it does so at a much faster rate than other approaches.
Figure~\ref{fig:rep} visualizes the final representation of the example image after 1.6 minutes, and the 3D mesh of the Thai Statue constructed with marching cubes after 30 minutes.
WIRE achieves the highest accuracy both for images (43.2dB) and for the occupancy volume (0.99), underlying our hypothesis that INRs equipped with a Gabor nonlinearity have higher approximation accuracy.

\subsection{Solving inverse problems of 2D images}
WIRE's inductive bias favors images, and hence can be used for solving linear inverse problems.
To demonstrate the advantages of WIRE as a strong prior for images, we showcase its performance on image denoising, single image super resolution, and multiimage super resolution.

\begin{figure*}[tt]
    \centering
    \includegraphics[width=\textwidth]{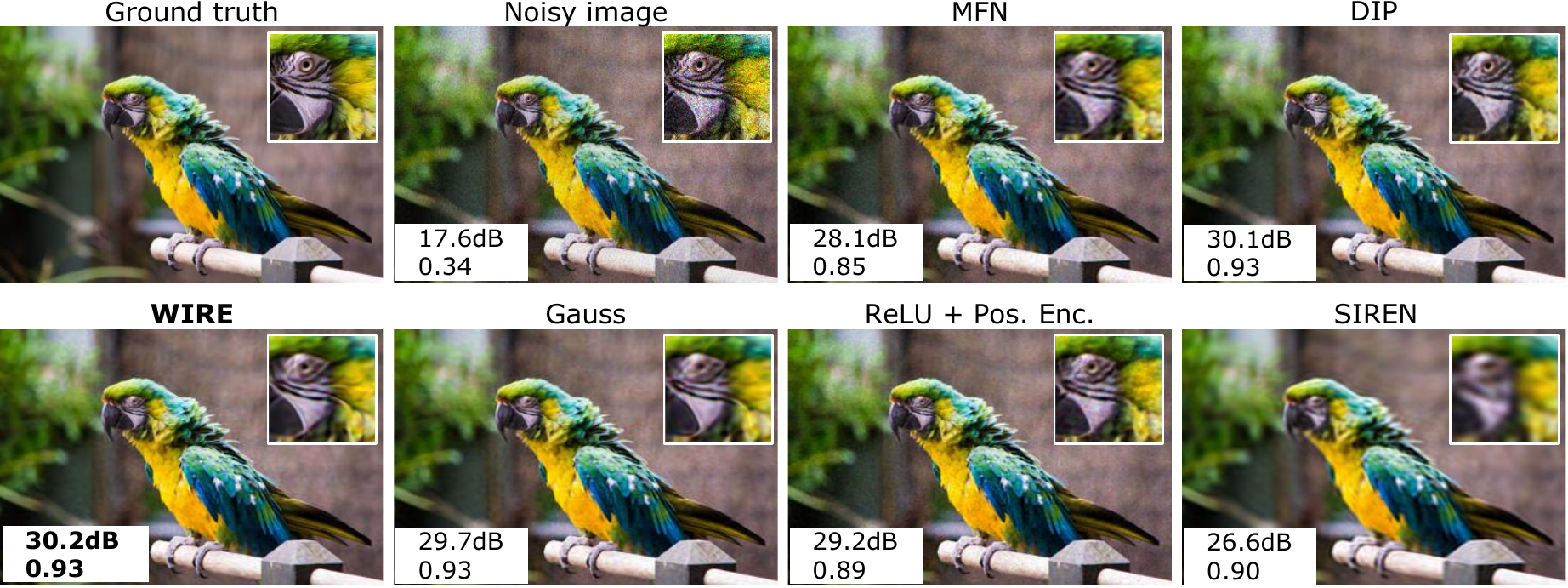}
    \caption{\textbf{WIRE is robust to noise.} A powerful feature uniquely enabled by WIRE is the robustness to noisy data. Here, we show an image representation with added shot noise, resulting in an input PSNR of 17.6dB. Among the various approaches, WIRE results in the highest PSNR and SSIM of any representation, thereby naturally resulting in denoising. }
    \label{fig:imdenoise}
\end{figure*}

\paragraph{Image denoising.} To evaluate the robustness of INRs for representing noisy signals, we learned a representation on a high resolution color image from the DIV2K dataset~\cite{Agustsson_2017_CVPR_Workshops}.
We simulated photon noise with an independently distributed Poisson random variable at each pixel with a maximum mean photon count of $30$, and a readout count of $2$, resulting in an input PSNR of 17.6 dB.
We then learned a representation on this noisy image with various nonlinearities.
In all cases, we chose an MLP with two hidden layers and 256 features per layer.
We also compared the denoising result with deep image prior (DIP)~\cite{ulyanov2018deep}.
Figure~\ref{fig:imdenoise} visualizes the final result for each nonlinearity along with metrices for each result.
WIRE produces the sharpest image with least amount of residual noise. 
Qualitatively, WIRE's result is similar to DIP's, implying that WIRE enjoys inductive biases that make it a good choice for inverse problems.

\begin{figure*}[tt]
    \centering
    \includegraphics[width=\textwidth]{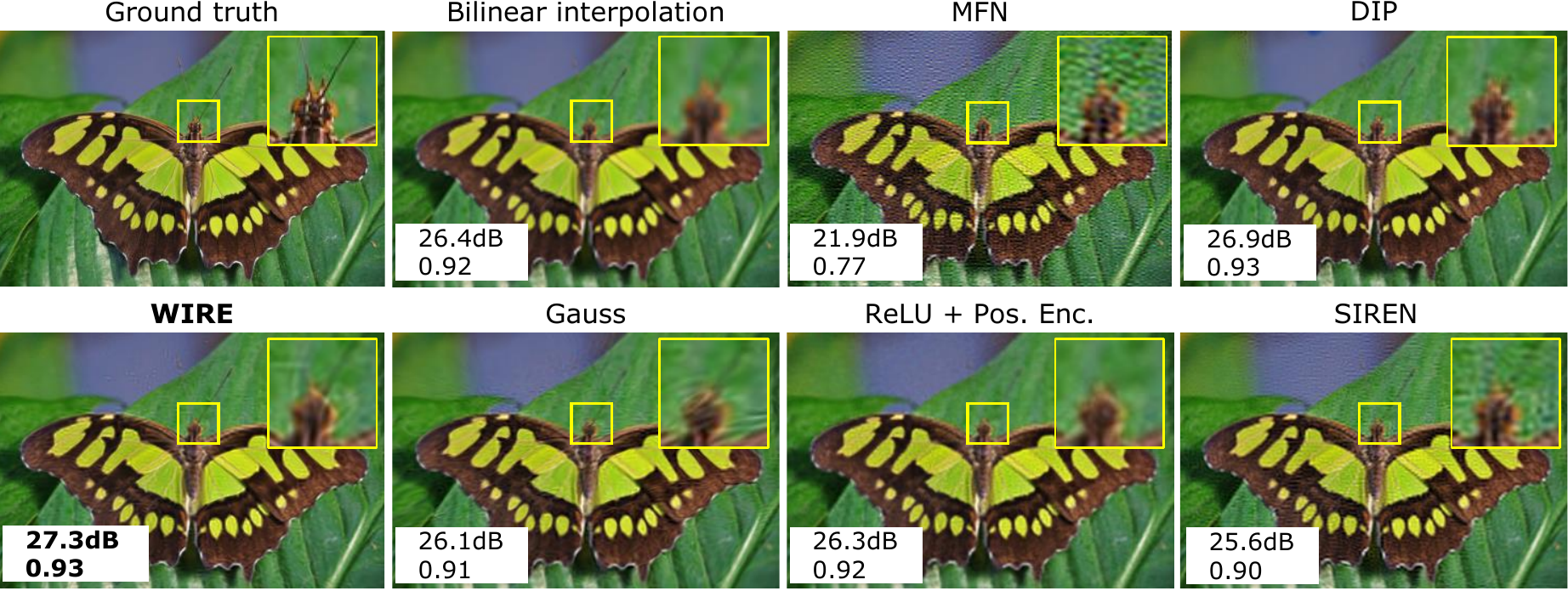}
    \caption{\textbf{WIRE for single image super resolution.} The figure above shows results for a $4\times$ single image super resolution with various approaches. Thanks to its strong implicit bias, WIRE results in the sharpest reconstruction with quantitatively higher reconstruction metrics.}
    \label{fig:sr}
\end{figure*}

\begin{figure*}[tt]
    \centering
    \includegraphics[width=\textwidth]{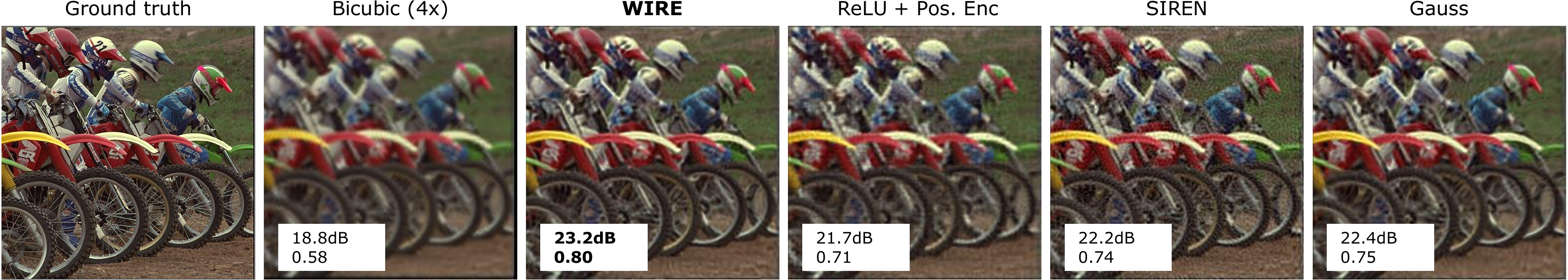}
    \caption{\textbf{Multi-image super resolution.} INRS are particularly appealing for handling data on an irregular grid, such as images captured with multiple sub-pixel shifts. The figure above shows $4\times$ super resolution with 4 images captured with varying sub-pixel shifts and rotations. We then solved a joint inverse problem where the high resolution image is modeled as the output of an INR. WIRE produces the best reconstruction both quantitatively and qualitatively, implying that WIRE has favorable interpolation properties for visual signals.}
    \label{fig:multi}
\end{figure*}

\paragraph{Image super resolution.}
INRs function as interpolatants, and hence super resolution is expected to benefit from INRs with good implicit biases.
We evaluate this hypothesis by implementing $4\times$ super resolution on a DIV2K image.
The forward operator can be cast as $\vy = A_4 \vx$ where $A_4$ implements a $4\times$ downsampling operator (without aliasing).
We then solved for the sharp image by modeling $\vx$ as output of an INR.
Figure~\ref{fig:sr} visualizes the result on super resolution of image of a butterfly with various approaches.
WIRE produces the sharpest result with crisp details on the butterfly's antenna and on the wings.
As with denoising, WIRE results are similar to DIP, establishing the ubiquity of WIRE.

INRs are particularly advantageous when data interpolation needs to be performed on an irregular grid.
An example of such settings is multi-image super resolution where the images are shifted and rotated with respect to each other.
Figure~\ref{fig:multi} shows an example of $4\times$ super resolution with four images (and hence $25\%$ compression) from the Kodak dataset~\cite{kodak1999} simulated with a small sub-pixel motion between them. 
The forward operator is then $\vy^k = A^k_4\vx$ where $A^k_4$ encodes the downsampling, and translation and rotation for the $k^\text{th}$ image. 
The visualizations in the figure demonstrate that WIRE achieves the highest accuracy and is qualitatively better at reconstructing high frequency components.
In contrast, the Gaussian nonlinearity leads to a blurry reconstruction, while SIREN results in ringing artifacts.
\begin{figure*}[!tt]
    \centering
    \includegraphics[width=0.95\textwidth]{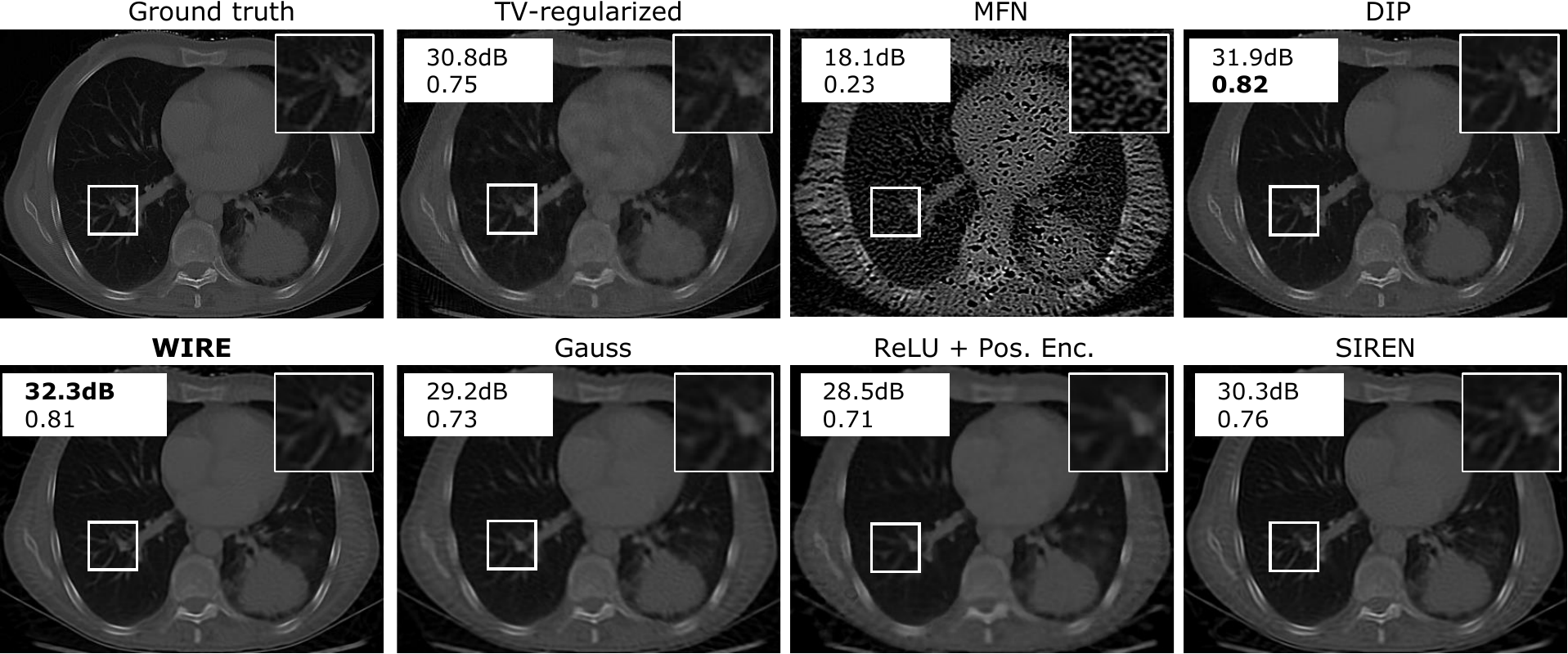}
    \caption{\textbf{Computed tomography reconstruction.} Inverse problems with noisy undersampled data require a strong signal prior for robust reconstruction. Here, we show CT-based reconstruction with 100 angles for a $256\times256$ image ($2.5\times$ compression) with various approaches. WIRE results in sharp reconstruction, exposing features that are blurry, or with ringing artifacts in reconstructions with other approaches. WIRE is hence a strong signal prior for images, and can solve a large class of inverse problems.}
    \label{fig:ct}
\end{figure*}

\begin{figure*}[!tt]
    \centering
    \includegraphics[width=0.98\textwidth]{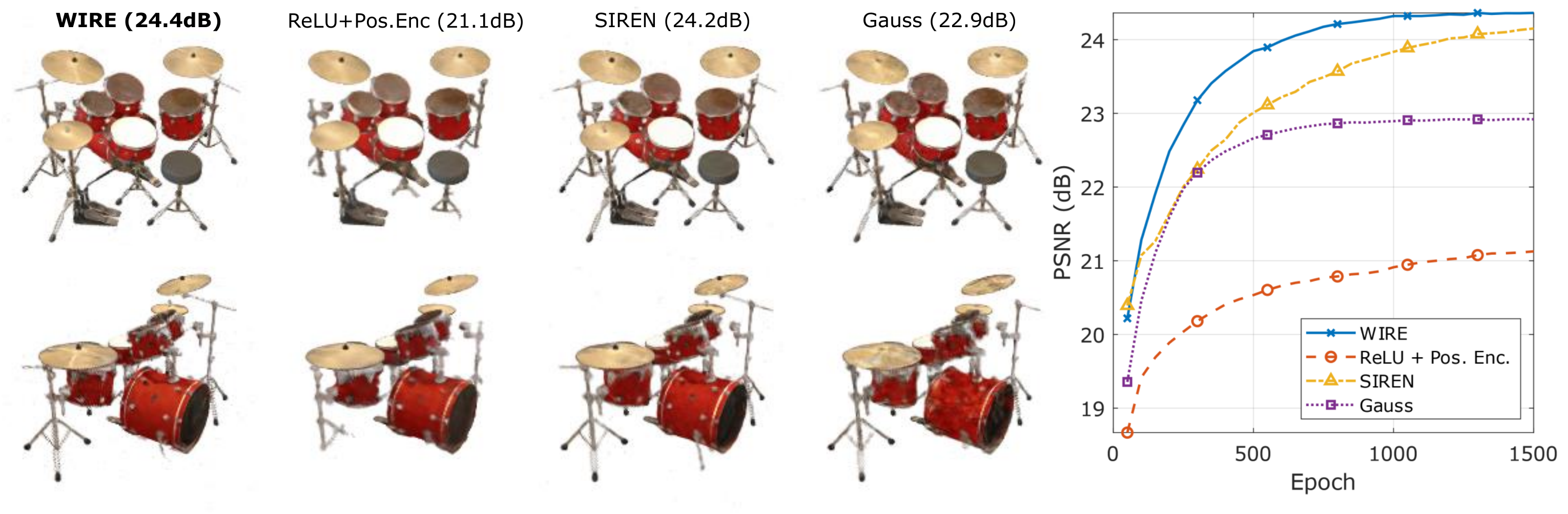}
    \caption{\textbf{Novel-view synthesis with neural radiance fields.} INRs have shown most promise in novel-view synthesis where the transmittance and color at each 3D voxel is modeled as output of INRs. Here, we show that WIRE is well-suited for novel-view synthesis with no additional positional encoding. WIRE not only achieves higher accuracy (+0.2dB) with fewer epochs, but captures details that are missed out by other nonlinearities, such as the rod connecting the ride cymbal to its stand and the anisotropic reflections on the cymbals.}
    \label{fig:nerf}
\end{figure*}
\paragraph{Computed tomography (CT) reconstruction.}
Strong signal priors are critical for solving underconstrained problems, and CT reconstruction is one such example.
%
%CT reconstruction is an example where signal priors play an important role.
%
We emulated 100 CT measurements of a $256\times256$ x-ray colorectal image~\cite{clark2013cancer}.
%
%We then added realistic shot noise equal to a maximum of 100 photons per sample to the measurements.
%
Figure~\ref{fig:ct} shows the final reconstruction with various approaches.
WIRE results in the sharpest reconstruction with clearly pronounced features.
SIREN performs the second best but has striation artifacts that are expected from an unregularized reconstruction.
The Gaussian nonlinearity results in overly smooth results.
WIRE can hence be used as a robust prior for inverse problems with noisy and undersampled measurements.

\subsection{Learning neural radiance fields}
INRs have been leveraged successfully for novel-view synthesis with neural radiance fields (NeRF)~\cite{mildenhall2020nerf}.
Given images from a sparse set of view points, the goal is to render an image from a different view point that is not in the training set.
NeRF achieves this by training a common INR that takes 3D location and viewing directions as inputs (hence a 5D input), and produces transmission and color at that location.
Images are then produced by integrating along lines that pass through each view's lens (pinhole). 
The simplest NeRF architecture consists of positional encoding, and two MLPs equipped with ReLU for transmission and color values.
We show that WIRE without any positional encoding produces higher quality results within fewer epochs.
\begin{figure}[!tt]
    \centering
    \includegraphics[width=\columnwidth]{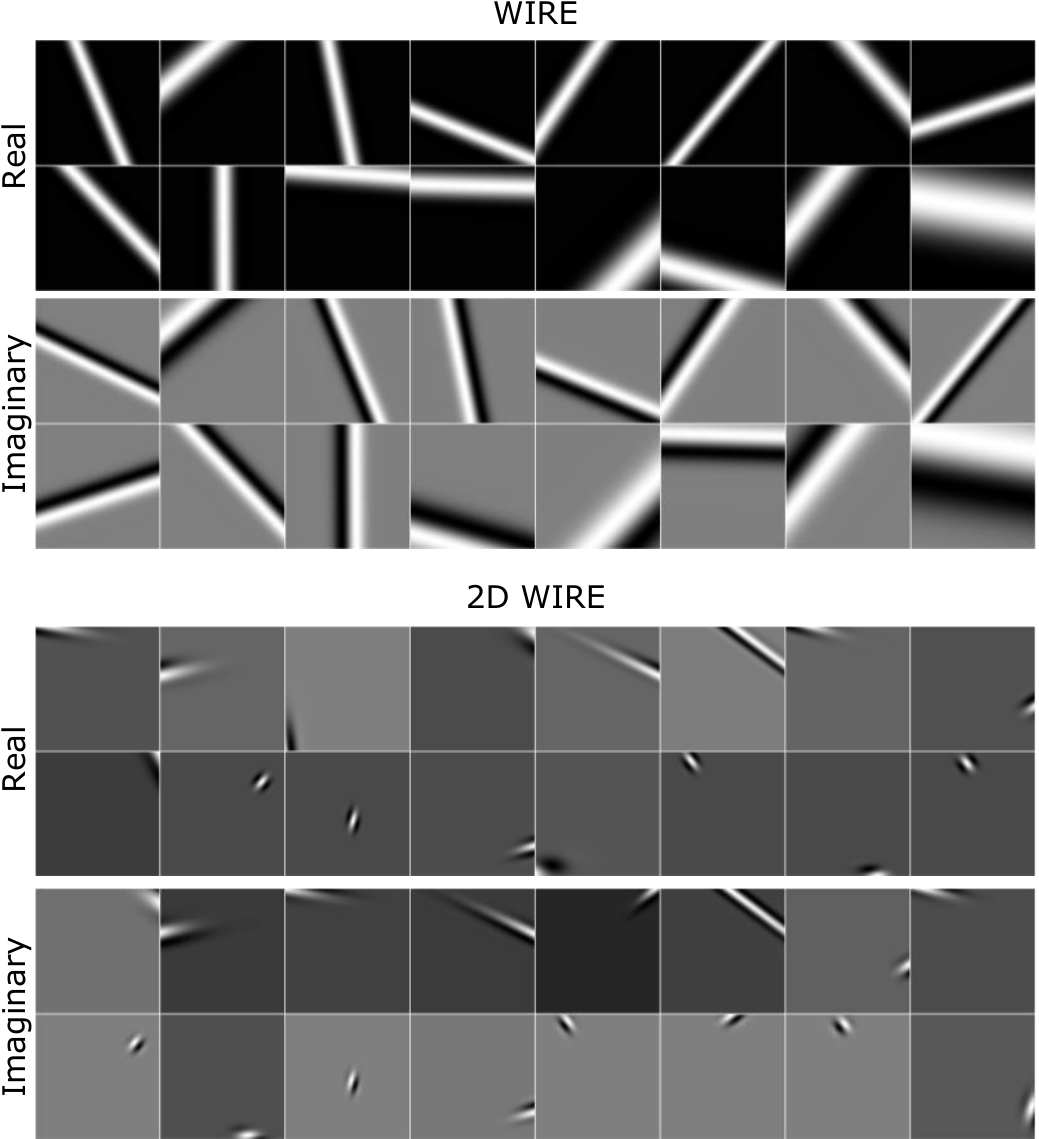}
    \caption{\textbf{First layer outputs with multi-dimensional WIRE.} The figure above shows outputs after first hidden layer with WIRE and 2D WIRE. We observe that 2D WIRE has spatially compact outputs due to the second Gaussian window, while WIRE has elongated structures orthogonal to the Gaussian window.}
    \label{fig:wire_2d_act}
\end{figure}
We trained NeRFs for reconstruction on the synthetic drum dataset~\cite{mildenhall2020nerf}.
Each image was downsampled to a resolution of $200\times200$.
To show the advantages of WIRE, we trained the radiance field with only 25 images instead of the default 100 images.
We used the ``torch-NGP" codebase~\cite{torch-ngp} for training the NeRF model.
For all experiments, we chose a 4-layered MLP with a width of 128 features for each layer.
Parameters and learning rate were chosen to achieve fastest rate of increase of approximation accuracy on the validation dataset.
Figure~\ref{fig:nerf} shows results with various nonlinearities.
WIRE produces highest accuracy (+0.2dB) with fastest rate of increase.
WIRE learns features absent in outputs of other nonlinearities such as the rod connecting the ride cymbal to its stand and the anisotropic reflections on the cymbal.

\subsection{Multi-dimensional WIRE comparisons}
\label{sec:multi-perf}

\begin{figure}[!tt]
    \centering
    \begin{subfigure}[t]{0.48\columnwidth}
        \centering
        \includegraphics[width=\columnwidth]{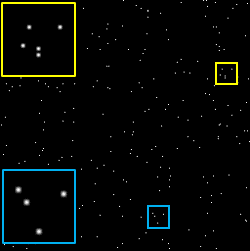}
        \caption{Ground truth}
    \end{subfigure}
    \begin{subfigure}[t]{0.48\columnwidth}
        \centering
        \includegraphics[width=\columnwidth]{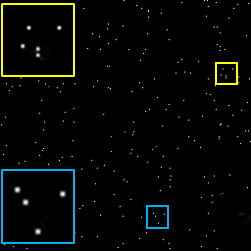}
        \caption{2D WIRE}
    \end{subfigure}
    \begin{subfigure}[t]{0.48\columnwidth}
        \centering
        \includegraphics[width=\columnwidth]{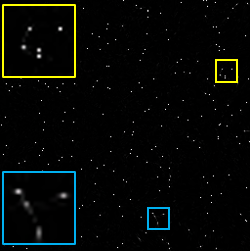}
        \caption{WIRE}
    \end{subfigure}
    \begin{subfigure}[t]{0.48\columnwidth}
        \centering
        \includegraphics[width=\columnwidth]{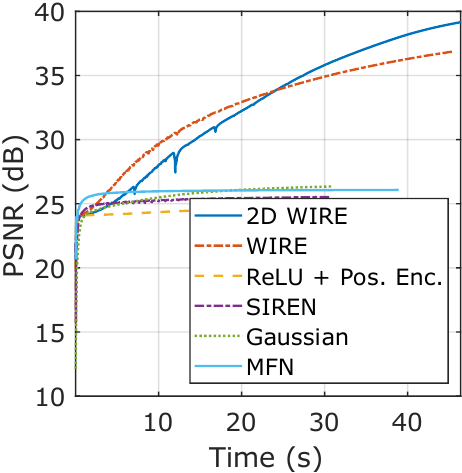}
        \caption{Accuracy vs. time}
    \end{subfigure}
    \caption{\textbf{Multi-dimensional localization.} The spatially compact nature of 2D WIRE enables representing sparse images. In this example, we show representation of a $256\times256$ image with 256 non-zero values. 2D WIRE represents each dot sharply while WIRE tends to blur the features along one of the two axes. Both WIRE and 2D WIRE converge rapidly compared to other approaches, as visualized in the plot in the bottom right corner.}
    \label{fig:wire_2d_img}
\end{figure}

As we discussed in Section~\ref{sec:multi}, WIRE can be instantiated as a multi-dimensional non-linearity.
One advantage of a two-dimensional WIRE is that its activations tend to be compact along both spatial axes (see Fig.~\ref{fig:wire_2d_act}).
This enables a more accurate fit for signals that are composed of spatially compact structures, such as many natural images.
Figure~\ref{fig:wire_2d_img} demonstrates the advantage of 2D WIRE's more general spatial localization for representing an image of point singularities.  
2D WIRE results in a sharper representation than 1D WIRE, which blurs out some of the points.

\begin{figure}[!tt]
    \centering
    \includegraphics[width=\columnwidth]{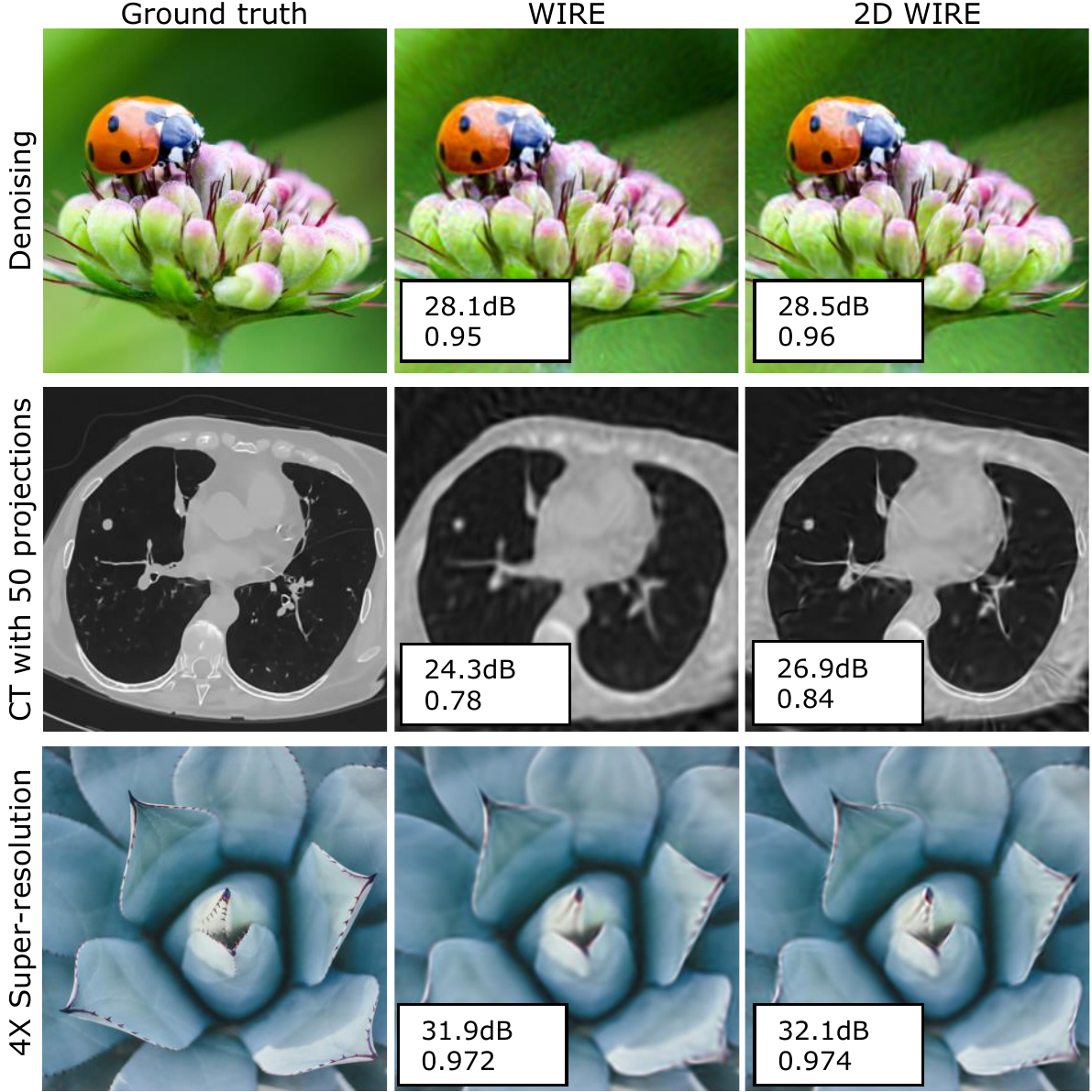}
    \caption{\textbf{Performance with multi-dimensional WIRE.} The figure above shows various linear inverse problems solved with INRs equipped with WIRE and 2D WIRE. Across the board, 2D WIRE achieves higher performance in terms of PSNR and SSIM. Visually, we observe that CT reconstruction and super resolution is significantly sharper with 2D WIRE owing to the compact nature of activations.}
    \label{fig:wire_2d_comp}
\end{figure}

The spatial localization of multi-dimensional WIRE is also of great advantage for solving inverse problems more robustly.
Figure~\ref{fig:wire_2d_comp} compares WIRE and 2D WIRE on denoising, CT reconstruction, and 4$\times$ super-resolution tasks.
In all tasks, $\omega_0$ and $s_0$ of the mother wavelet were chosen for best performance, and the total number of parameters were the same for WIRE and 2D WIRE. 
Across the board, 2D WIRE achieves higher accuracy than WIRE. 
2D WIRE also learns sharper features for the CT reconstruction of lungs while WIRE's tend to be more blurry. 
Similarly, in the super-resolution task, 2D WIRE learns improved high frequency features to represent the center of the flower.

%% file: conclusions.tex
We have proposed and validated the advantages of WIRE that equips INRs with a complex Gabor wavelet activation nonlinearity. 
We have shown with an extensive set of experiments that WIRE (a) has higher representation capacity, (b) achieves higher accuracy at a faster rate, and (c) has strong inductive biases that make it compelling for solving challenging inverse problems.
Other activation nonlinearities have largely complementary strengths: SIREN has high representation capacity and trains fast, but is poor at regularizing inverse problems. Positional encoding has lower capacity but is a good choice for novel-view synthesis. Gaussian nonlinearity is more favorable for denoising tasks.
However, WIRE inherits the best properties of all of the above activation nonlinearities, and hence is the current go-to INR solution for signal representation and solving inverse problems.

%
%\daniel{in various tasks, other INRs change order of what is best---SIREN for image representation, Pos. Enc. + ReLU for 3D, Gauss for denoising---but WIRE is consistently the best, inheriting the best properties of each.}

%% file: ack.tex
This work was supported by NSF grants CCF-1911094, IIS-1838177, and IIS-1730574; ONR grants N00014-18-12571, N00014-20-1-2534, and MURI N00014-20-1-2787; AFOSR grant FA9550-22-1-0060; and a Vannevar Bush Faculty Fellowship, ONR grant N00014-18-1-2047.

%% file: appendix1.tex
\subsection{WIRE initialization}
\begin{figure}[!tt]
    \centering
    \includegraphics[width=\columnwidth]{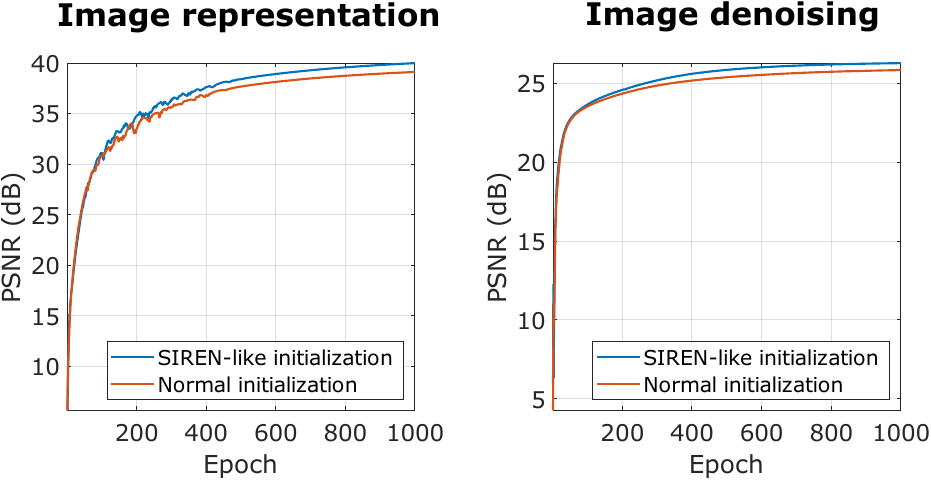}
    \caption{\textbf{Effect of initialization.} The plots show approximation accuracy for image representation and image denoising (20dB input PSNR) across training epochs with SIREN-like weight initialization and standard initialization. WIRE is robust to the initial weights, but marginally benefits from a SIREN-like initialization.}
    \label{fig:init}
\end{figure}

INRs like SIREN~\cite{sitzmann2020implicit} strongly depend on initialization to obtain accurate representation.
WIRE does not require any initialization except for the default uniform weights.
However, since WIRE consists of a complex sinusoidal term, it marginally benefits from SIREN-like initialization.
To understand the dependence, we evaluated approximation accuracy for image representation (no noise), and image denoising (20dB image noise).
Here, a SIREN-like weight initialization implies the first layer weights are drawn from $\mathcal{U}(-1/N, 1/N)$ and the weights of the rest of the layers are drawn from $\mathcal{U}(-\sqrt{6/(\omega_0N)}, \sqrt{6/(\omega_0N)})$, where $N$ is the number of input features and $\mathcal{U}(a, b)$ is a uniform distribution over $[a, b]$.
A normal weight initialization involves drawing weights from $\mathcal{U}(-1/\sqrt{N}, 1/\sqrt{N})$ for all layers.
Fig.~\ref{fig:init} compares the representation accuracy for SIREN-like and standard initialization.
%(\vishwa{I think normal is not the right word?}).
%
In both cases, we see that the trends are nearly similar; SIREN-like initialization results in up to 1dB higher accuracy.
Hence WIRE is largely robust to initial parameters which enables easy tuning across a large range of hyperparameters $\omega_0, s_0$.
\begin{figure}[!tt]
    \centering
    \includegraphics[width=\columnwidth]{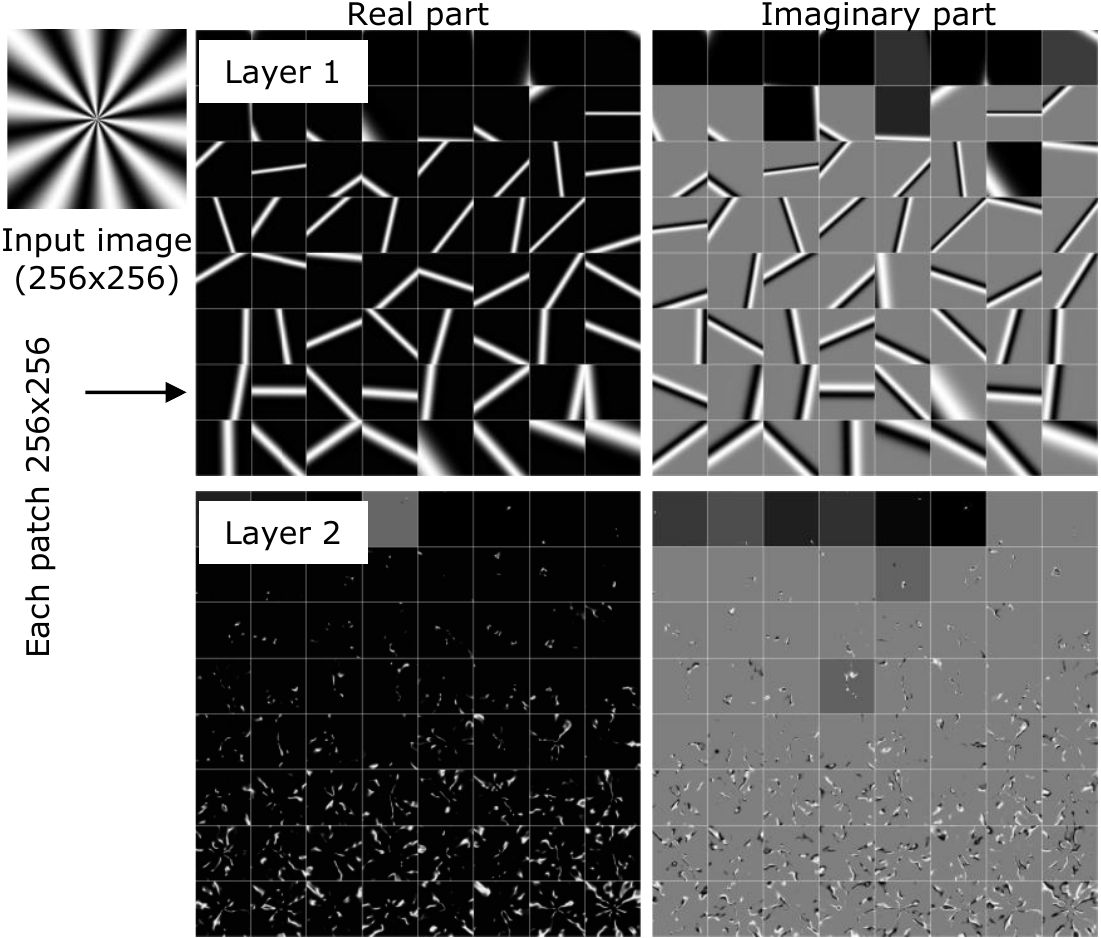}
    \caption{\textbf{Layer outputs for WIRE.} The input image is a Siemens star test image that contains all spatial frequencies and all angles. The patches show outputs (same size as image) of each hidden feature in layer one and two for a two hidden-layer MLP equipped with WIRE nonlinearity. WIRE results in sparse images which enables high representational capacity for images, as shown in Fig.~\ref{fig:layervis}.}
    \label{fig:wirevis}
\end{figure}
\begin{figure*}[!tt]
    \centering
    \includegraphics[width=\textwidth]{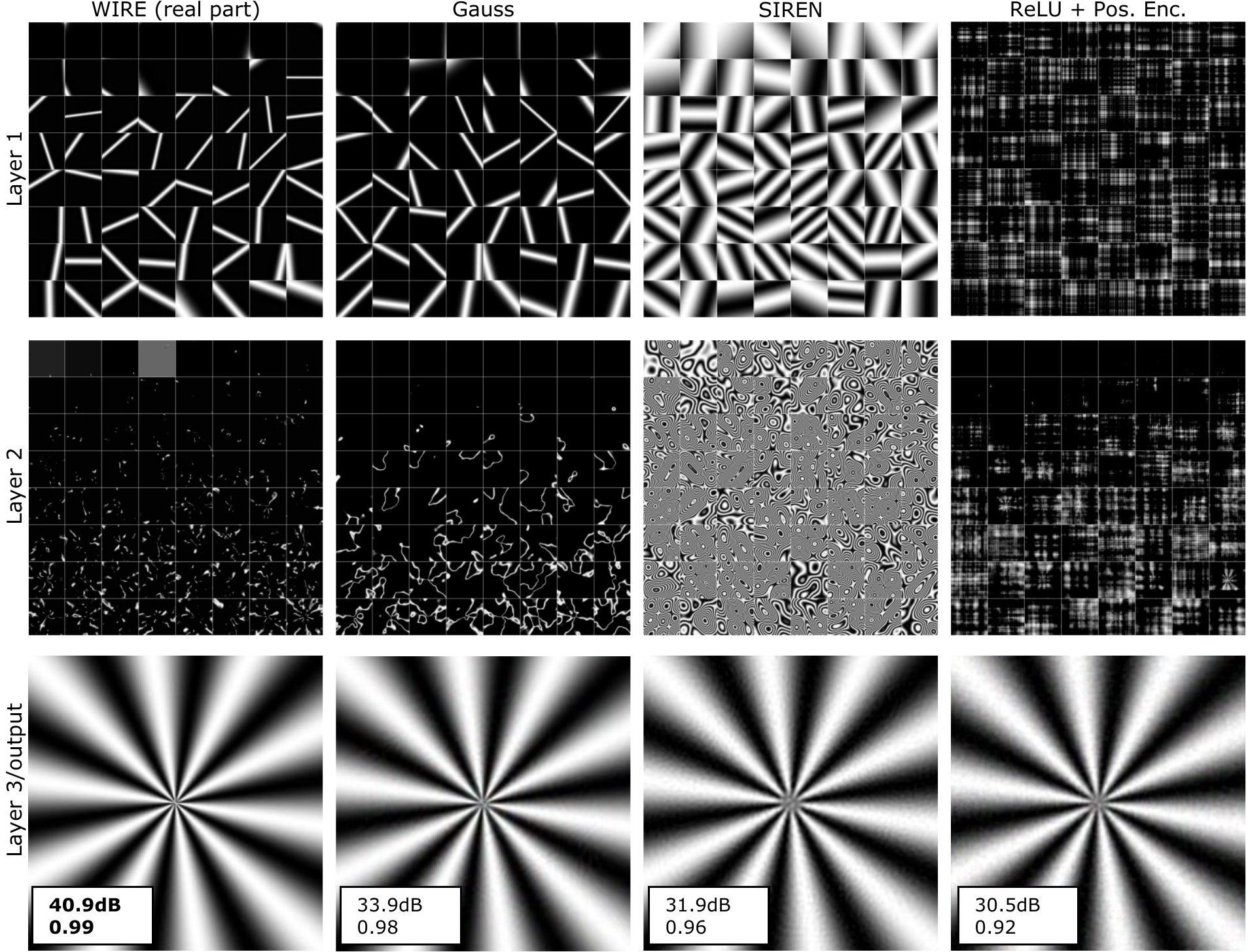}
    \caption{\textbf{Visualization of hidden layer outputs.} The figure above visualizes outputs of hidden features in the two layers for the Siemens sector test image shown in Fig.~\ref{fig:wirevis}.
    WIRE uniquely results in sparse images, which enables high accurate representation of high frequency parts of the image (center of the sector).}
    \label{fig:layervis}
\end{figure*}
\subsection{WIRE layer visualizations}
Gabor wavelets uniquely enable space--frequency localization, a property we observe is inherited by WIRE. %(\vishwa{what is the best way to say this?}).
To evaluate this hypothesis, we visualized the output WIRE composed of an MLP with two hidden layers and 181 hidden features each.
We then learned a representation for a Siemens star test image that consists of all spatial frequencies and orientations.
Fig.~\ref{fig:wirevis} visualizes the input image real and imaginary outputs of 64 hidden features with least variance.
The output of each first layer feature consists of one-dimensional Gabor wavelets at various orientations, while the outputs of second layer consist of sparsely populated images.

Fig.~\ref{fig:layervis} visualizes outputs at each layer for various nonlinearities and the final approximated image for the Siemens star test image.
The sparse outputs of second layer are evidently unique to WIRE.
Gauss has outputs that look less spares, while SIREN and ReLU with positional encoding result in dense outputs.
This has a direct consequence on approximation capacity for high frequency parts of the signal.
The final result in the bottom row shows that the sparse nature of outputs of WIRE enables high approximation accuracy with qualitatively better features at the center of the image which consists of highest spatial frequenices.
Gauss follows next as it results in the second most sparse outputs at each layer.
SIREN and ReLU with positional encoding alike produce blurry outputs at the center, primarily due to the non-compact nature of outputs.
WIRE's ability to decompose images as a linear combination of sparse images results in high representational capacity for the same number of parameters, as we verify empirically in the next section.

\subsection{Sensitivity to training parameters}
WIRE is a promising INR model that achieves high representation accuracy and is robust to a wide range of training parameters.
We demonstrate the efficacy of WIRE in this section with several sensitivity analyses.

\begin{figure}[!tt]
    \centering
    \includegraphics[width=0.8\columnwidth]{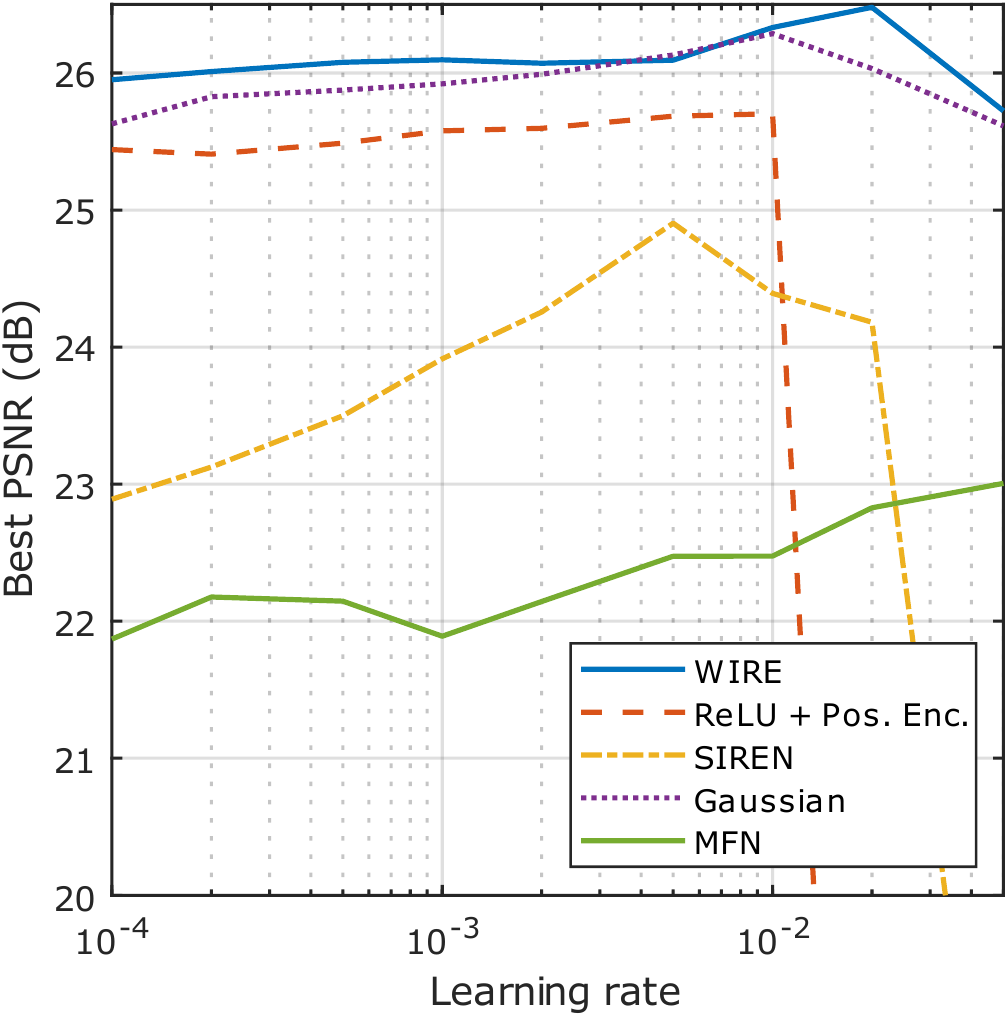}
    \caption{\textbf{Effect of learning rate.} The plot above shows approximation accuracy for representing a noisy image (input PSNR of 20dB) with various nonlinearities. WIRE is robust to learning rate, and produces best results with high learning rate of $2\times10^{-2}$.}
    \label{fig:lr}
\end{figure}

\paragraph{Effect of learning rate.}
WIRE performs well for a large range of learning rates.
To understand the performance trends, we learned an image representation with added noise (20dB input PNSR) with various nonlinearities.
We used a 2-layer MLP with 256 hidden features per layer.
Fig.~\ref{fig:lr} shows the maximum representation PSNR with varying learning rates.
WIRE has a stable and significantly higher accuracy compared to other approaches.
Interestingly, the highest accuracy is achieved at a high learning rate of $2\times10^{-2}$.
This behavior is also observed with deep image prior~\cite{ulyanov2018deep} where a larger learning rate enabled stronger regularization.
Such a similar behavior implies WIRE enjoys strong inductive biases and hence is amenable to solve inverse problems.

\begin{figure}[!tt]
    \centering
    \begin{subfigure}[t]{0.48\columnwidth}
        \includegraphics[width=\columnwidth]{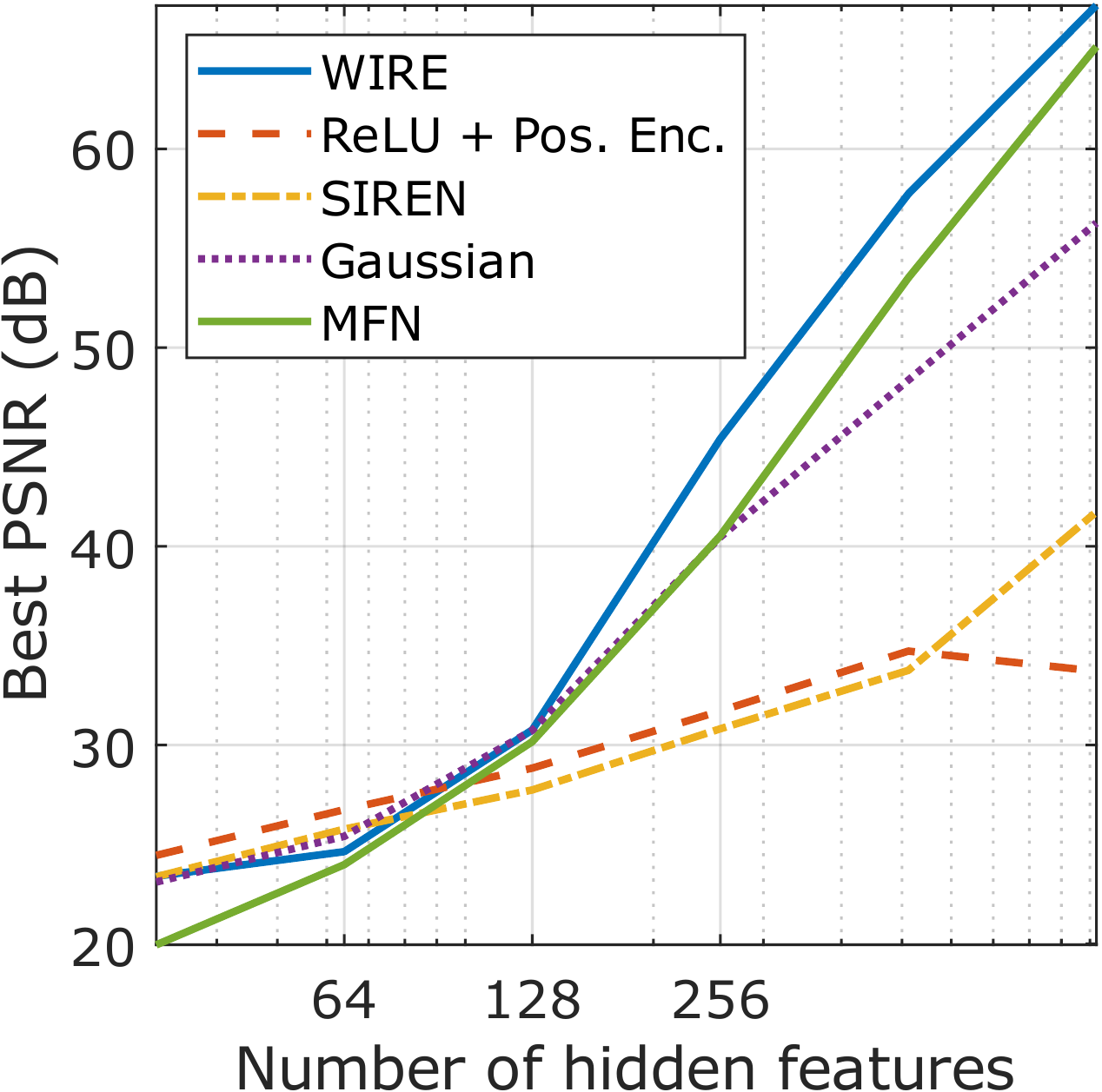}
        \caption{\centering PSNR vs.\ number of features}
        \label{fig:nlayers}
    \end{subfigure}
    \begin{subfigure}[t]{0.48\columnwidth}
        \includegraphics[width=\columnwidth]{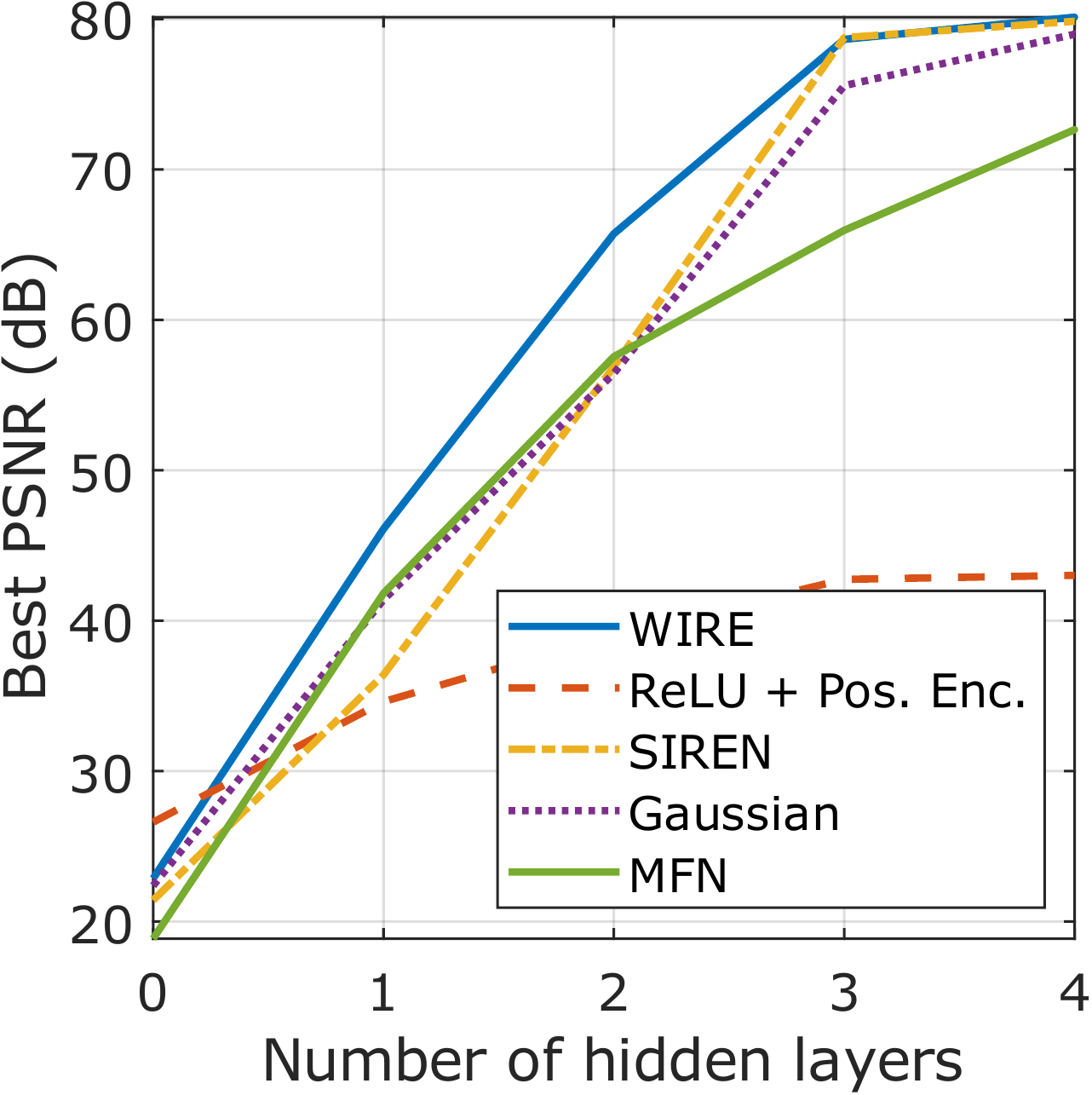}
        \caption{\centering PSNR vs.\ number of layers}
        \label{fig:nfeat}
    \end{subfigure}
    \caption{\textbf{Effect of number of parameters.} The plot above shows approximation accuracy for representing an image with varying number of (a) hidden features and (b) hidden layers. WIRE outperforms other nonlinearities with 128 or more hidden features, and one or more layers and is nearly the same as SIREN and Gaussian nonlinearities for more than 3 layers.}
\end{figure}

\paragraph{Effect of number of layers.}
Fig.~\ref{fig:nlayers} shows a plot of representation accuracy of an image for varying number of hidden layers with various nonlinearities.
In each case, the number of hidden features were set to 256.
We reduced the learning rate with increasing layers to avoid divergence.
WIRE uniformly outperforms other approaches (except with 0 hidden features), as is to be expected as Gabor wavelets enable high approximation accuracies for images.
Interestingly, for a large number of hidden layers ($\ge 3$), WIRE performance is similar to SIREN and Gaussian nonlinearity.
This is to be expected as the network has a large capacity with so many layers.
However, a large number of layers is computationally expensive and often results in an unstable learning regime.
WIRE therefore is a reliable choice for small to medium number of hidden layers for most cases.

\begin{figure}[!tt]
    \centering
    \begin{subfigure}[t]{0.4\columnwidth}
        \includegraphics[width=\columnwidth]{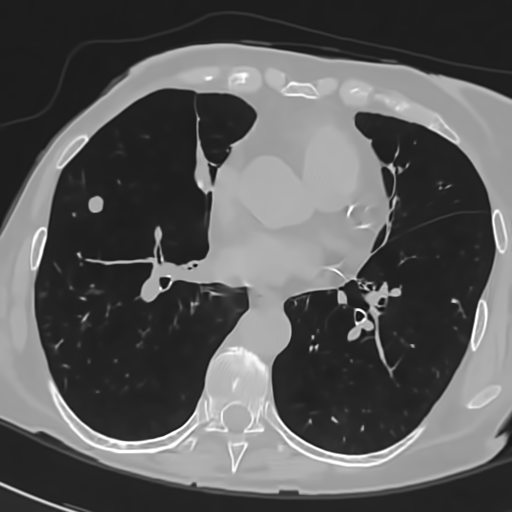}
        \caption{\centering Ground truth image}
        \label{fig:ct_gt}
    \end{subfigure}
    \begin{subfigure}[t]{0.58\columnwidth}
        \includegraphics[width=\columnwidth]{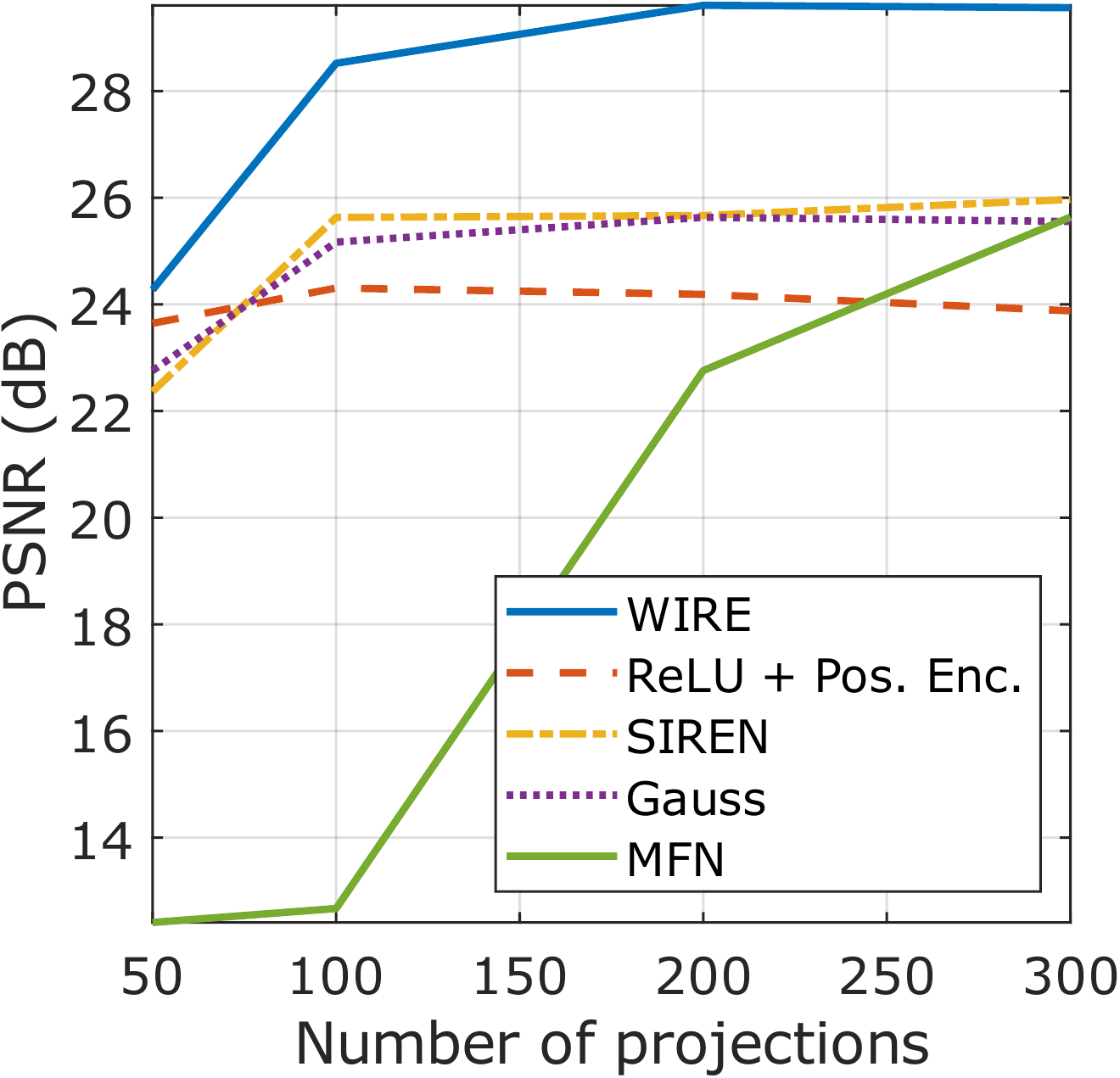}
        \caption{\centering PSNR vs.\ number of prjections}
        \label{fig:ct_sweep}
    \end{subfigure}
    \caption{\textbf{CT with varying number of projections.} (a) shows the $512\times512$ ground truth x-ray image of lungs~\cite{armato2011lung} we used for our CT experiments. We denoised the original image to remove streak artifacts. (b) shows accuracy as a function of number of measurements with various nonlinearities. Across the board, WIRE outperforms all other approaches by a considerable margin.}
    \label{fig:ct1}
\end{figure}

\paragraph{Effect of number of features.}
Fig.~\ref{fig:nfeat} shows approximation accuracy for image representation with varying number of hidden features.
In all cases, the number of hidden layers were fixed to be two.
The performance of WIRE is similar to other nonlinearities at very low number of hidden features, where all models similarly lack sufficient richness.
%(\vishwa{Is there a reason?}). 
%
For higher than 128 features, WIRE outperforms other approaches with MFN~\cite{fathony2020multiplicative} coming a close second.

\begin{figure*}[!tt]
    \centering
    \includegraphics[width=\textwidth]{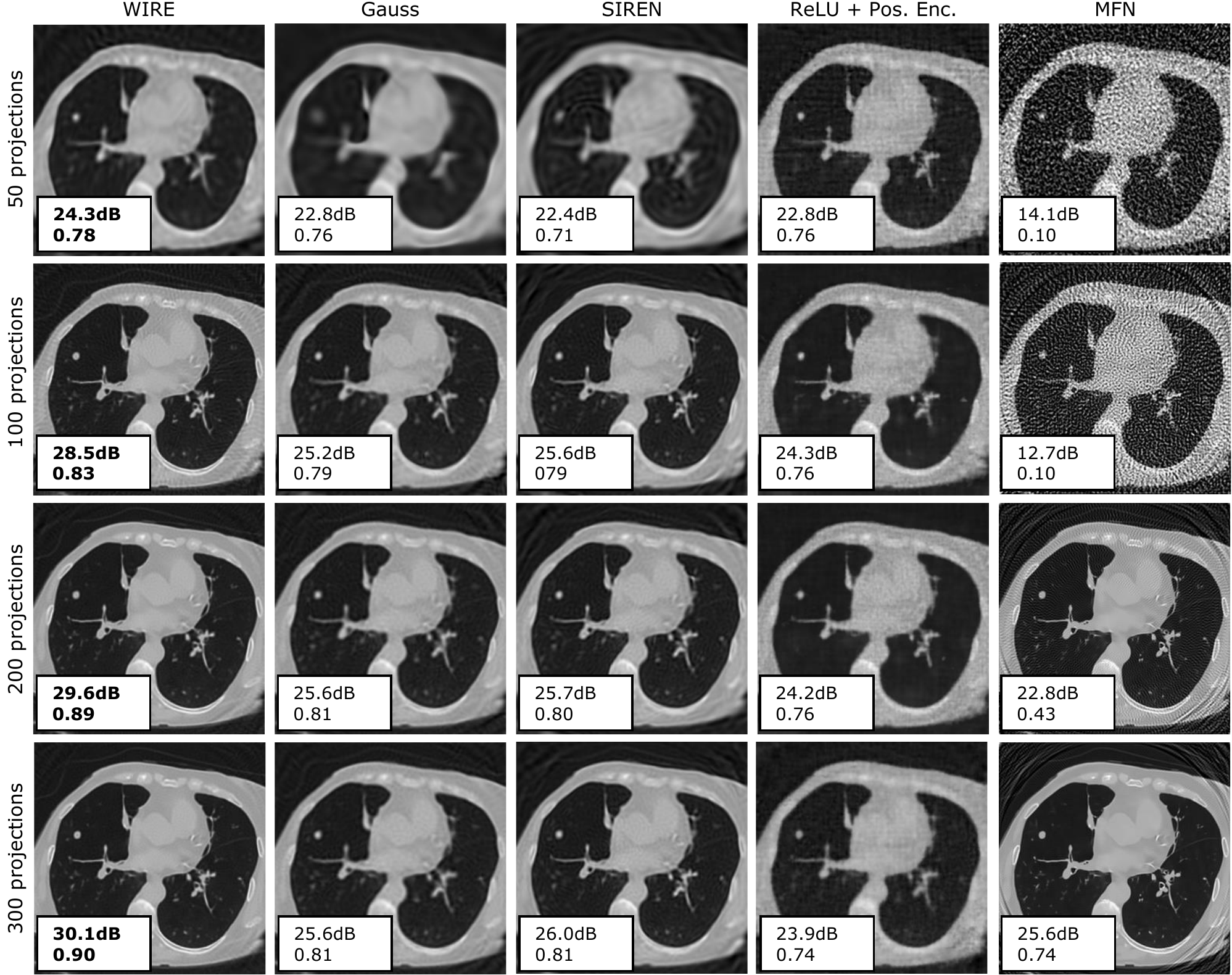}
    \caption{\textbf{Effect of number of projections on CT accuracy.} The images above visualize reconstruction for the lungs image shown in Fig.~\ref{fig:ct_gt}. WIRE outperforms all other approaches even with 50 projections ($10\%$ measurements) and is visually pleasing.}
    \label{fig:ct_vis}
\end{figure*}

\begin{figure*}[!tt]
    \centering
    \includegraphics[width=\textwidth]{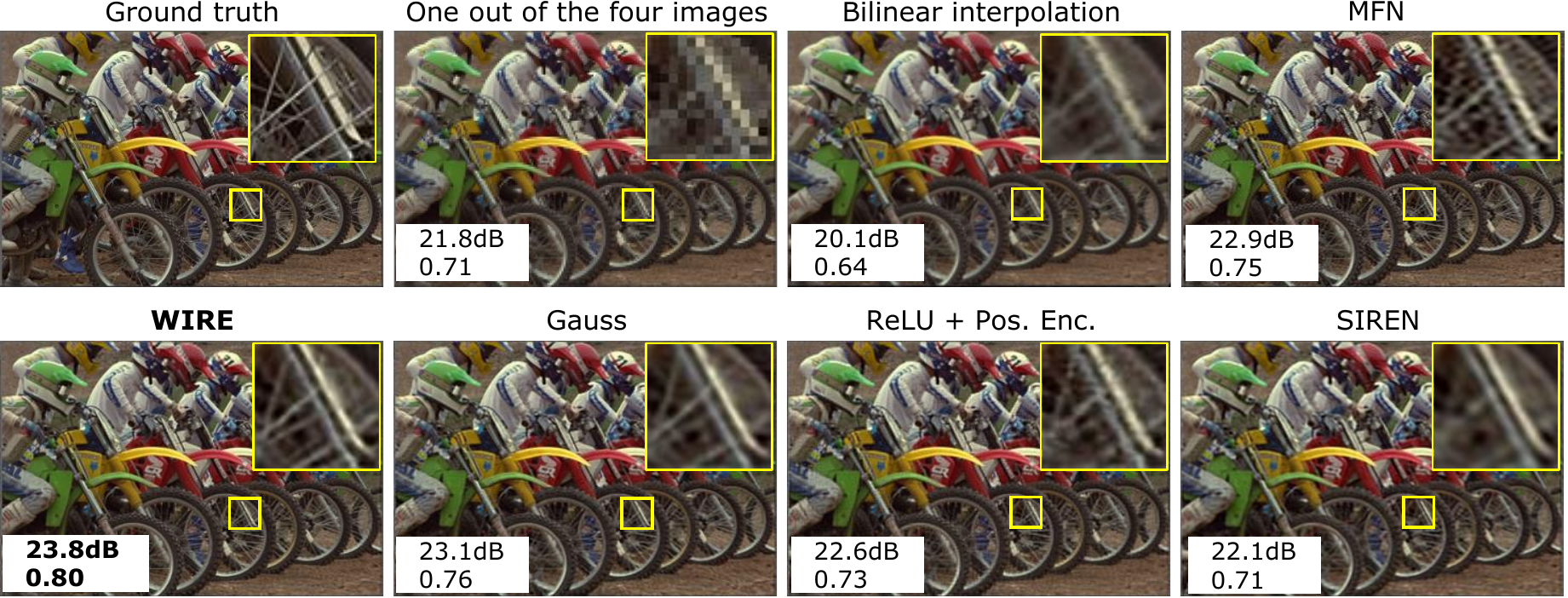}
    \caption{\textbf{Multi-image super-resolution.} The figure above visualizes multi-frame super-resolution where each sub-frame was simulated with a small known sub-pixel shift. WIRE achieves highest reconstruction accuracy with qualitatively better reconstruction.}
    \label{fig:multi}
\end{figure*}

\begin{figure}[!tt]
    \centering
    \begin{subfigure}[t]{0.48\columnwidth}
        \centering
        \includegraphics[width=\textwidth]{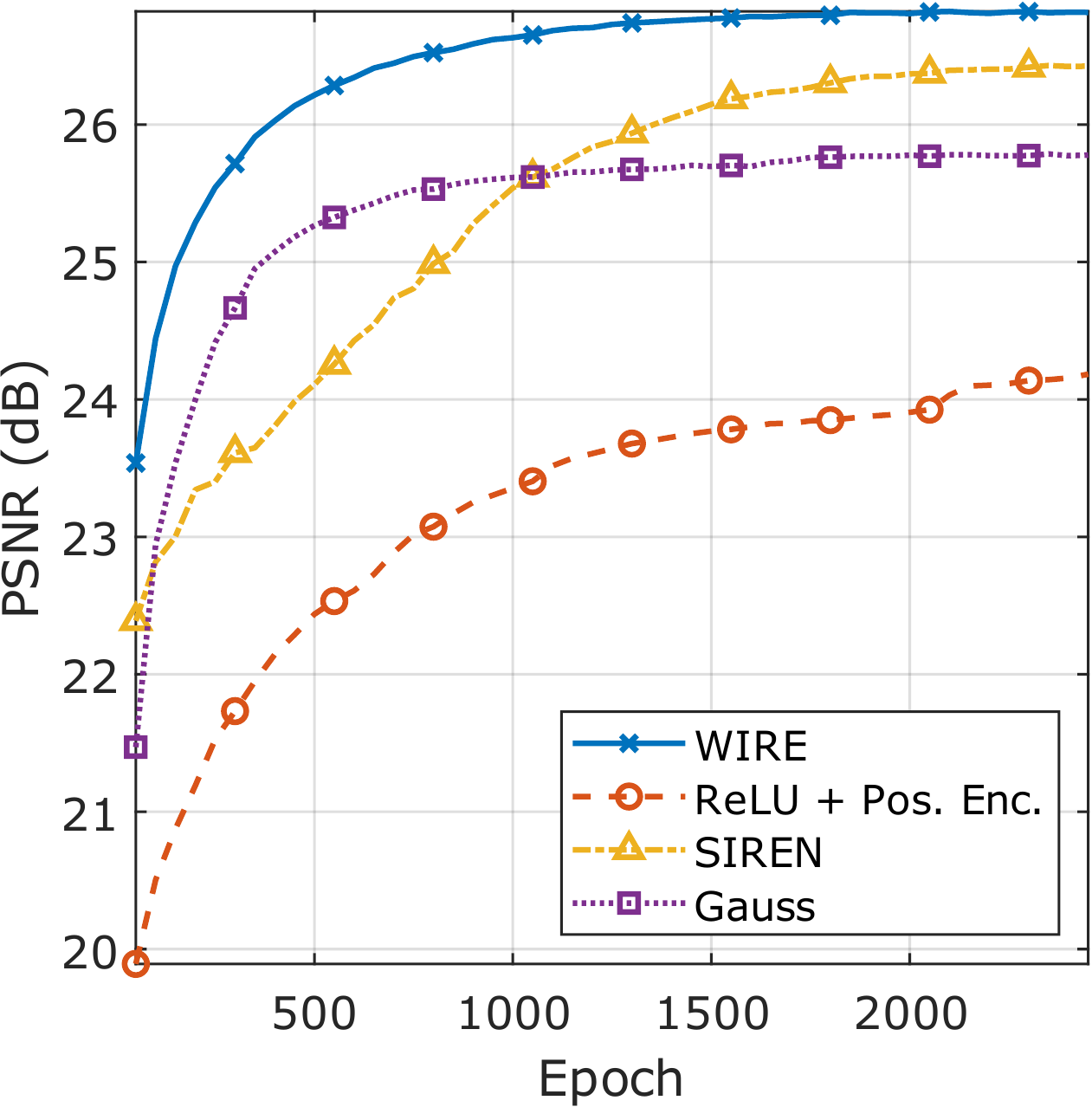}
        \caption{\centering Accuracy vs. epochs for 100 images.}
        \label{fig:drums100}
    \end{subfigure}
    \begin{subfigure}[t]{0.49\columnwidth}
        \centering
        \includegraphics[width=\textwidth]{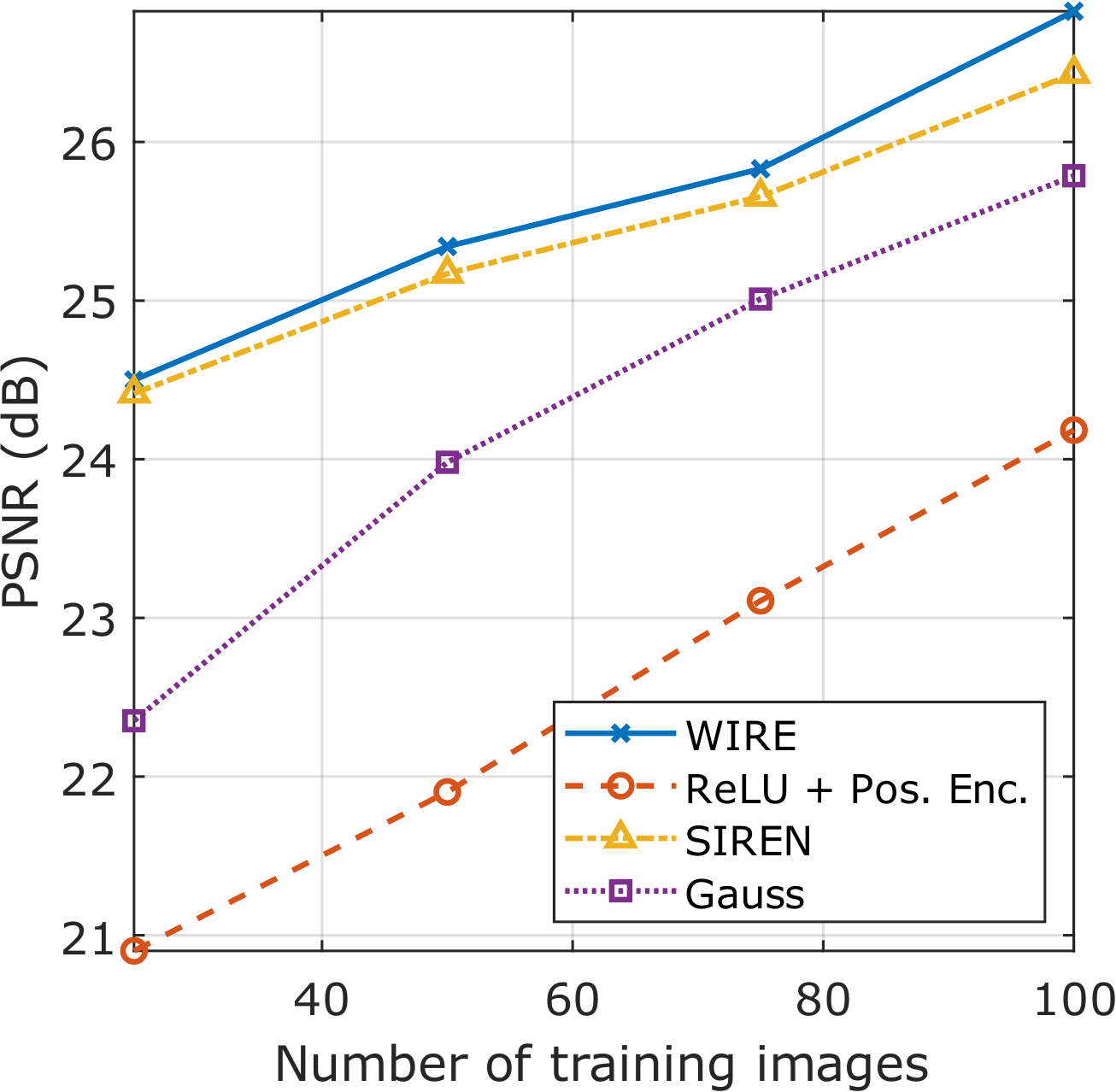}
        \caption{\centering Accuracy vs. number of images.}
        \label{fig:drums_nimg}
    \end{subfigure}
    \caption{\textbf{NeRF accuracy on drums dataset.} (a) shows training accuracy at each epoch for various nonlinearities with neural radiance fields when trained with 100 images. WIRE achieves $0.4$dB higher than the next highest (SIREN) when trained with 100 images and does so in a rapid manner. (b) shows accuracy as a function of number of images with WIRE outperforming other approaches for all number of images.}
\end{figure}

\begin{figure*}[!tt]
    \centering
    \includegraphics[width=\textwidth]{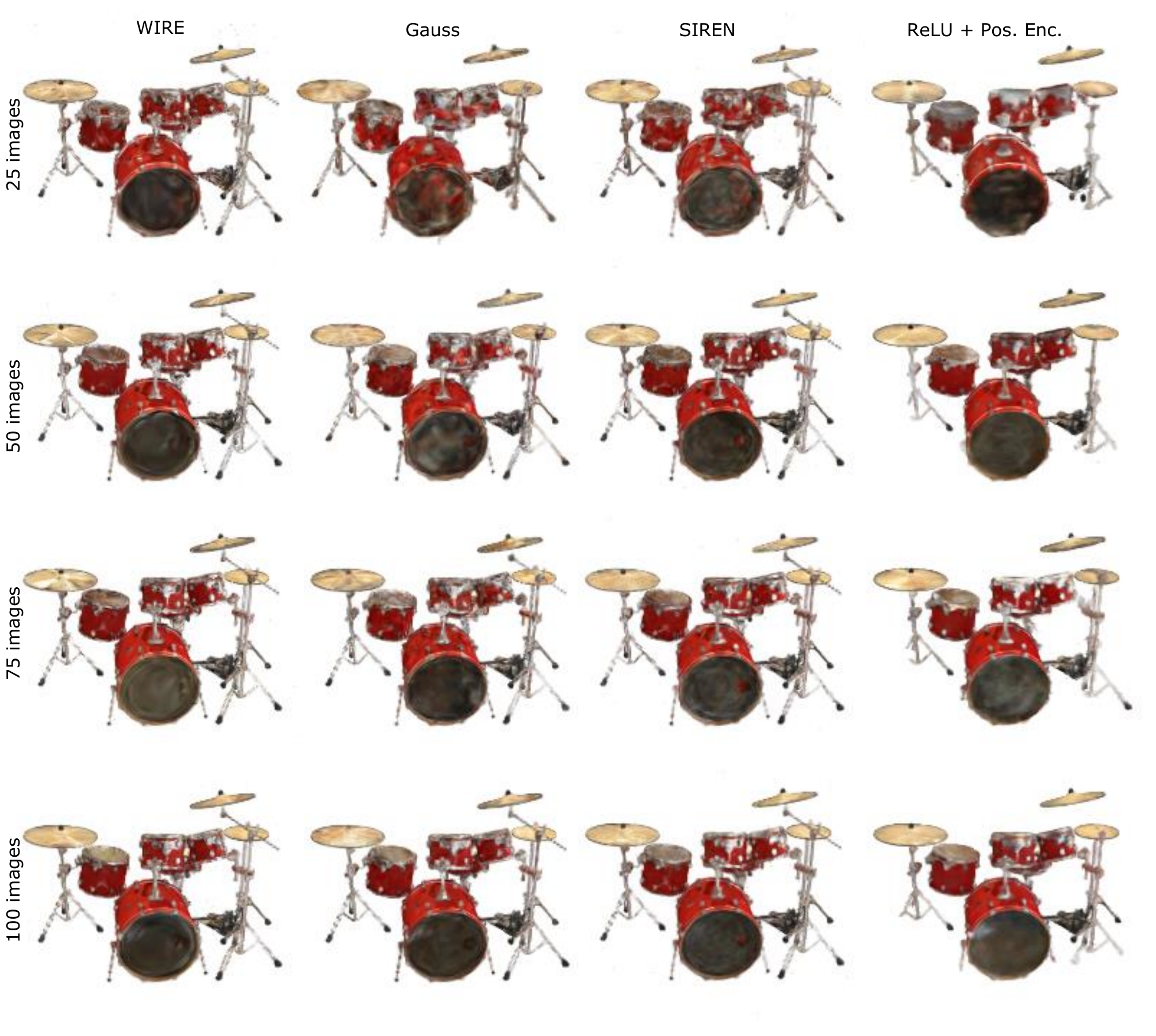}
    \caption{\textbf{Effect of number of images on NeRF accuracy.} The figure above visualizes rendered images with a neural radiance field for various nonlinearities (across columns) and varying number of images (across rows). WIRE achieves visually better reconstruction than all other methods for all numbers of images, thanks to its strong inductive biases that enable learning the high frequency features of the scene's radiance field.}
    \label{fig:drums}
\end{figure*}

\subsection{Inverse problems}
\paragraph{Computed tomographic reconstruction.} We showed in Section~\ref{sec:exp} that computed tomography (CT) benefits from inductive biases of INRs. Here, we study the effect of number of measurements.
Fig.~\ref{fig:ct_gt} shows the ground truth image we used in our experiments. 
We denoised the original $512\times512$ image~\cite{armato2011lung} with BM3D~\cite{dabov2007color} ($\sigma=0.1$) to remove streak artifacts.
We then simulated CT reconstruction with varying numbers of projections.
%
%Figure~\ref{fig:ct_sweep} visualizes reconstruction of CT image of lungs~\cite{armato2011lung} with variable number of projections with various nonlinearities.
%
In each case, we used an MLP with three hidden layers and 256 hidden features per layer.
We sampled the INR on a regular grid to first generate the image, and then use Radon transform to obtain the sinogram.
From the accuracy plot in Fig.~\ref{fig:ct_sweep}, we see that WIRE achieves higher PSNR than any other nonlinearity.
Fig.~\ref{fig:ct_vis} visualizes the reconstruction with varying number of projections for each nonlinearity.
The reconstruction is visually superior even with small number of projections, which is particularly beneficial for reducing exposure to x-rays during capture.

\paragraph{Multi-image super-resolution.} We showed a result on multi-image super-resolution in Section~\ref{sec:exp}.
Here, we provide more details about the experiment.
Figure~\ref{fig:multi} shows the $512\times768$ dimensional ground truth image from the Kodak dataset~\cite{kodak1999}.
We simulated a total of four low-resolution images by modifying each $4\times$ downsampled image by a small translation and rotation, thereby resulting in sub-pixel motion between the frames.
We assumed the transformation $A_k$ between the high-resolution frame $\vx$ and each low-resolution frame $\vy_k$ was known.
We represented the high resolution $\vx$ as output of an INR. In each case, the INR had three hidden layers with 256 hidden features.
We then solved a linear inverse problem to estimate the high resolution image.
Figure~\ref{fig:multi} shows the reconstructed output for each nonlinearity and their metrics.
The inset shows reconstruction of spokes in the motorcycle. 
Visually, WIRE generates the sharpest features without any ringing artifacts.
Moreover, WIRE results in 1dB or better reconstruction accuracy, and $0.04$ higher SSIM.

\subsection{Neural radiance fields}
\paragraph{Implementation details.} For all experiments, we used the \verb|torch-ngp| package~\cite{torch-ngp} that implements a wide variety of approaches for training neural radiance fields.
The architecture consists of two networks that predict transmittance (sigma) and the color at each voxel respectively.
Each of the two networks consisted of an MLP with four hidden layers and 182 hidden features each.
The color MLP took position (x,y,z) and direction ($\theta, \phi$) as inputs, while the transmittance MLP took only the position as input.
As with all other experiments, we used $182/\sqrt{2} = 128$ hidden features for WIRE to account for parameter doubling due to complex weights.
We downsampled the images by $4\times$ to ensure that the model and training data fit in the graphical processing unit's (GPU) memory.
In the results shown in Fig.~\ref{fig:nerf}, we used a total of 25 randomly chosen images to train the NeRF, and then validated it on 100 images.
We used a learning rate of $4\times10^{-4}$ for WIRE and $2.5\times10^{-4}$ for all other nonlinearities and reduced it to $0.1\times$ initial value over a total of 2500 training epochs.
Except for ReLU, we did not use any form of positional encoding with other nonlinearities as we wished to demonstrate the capacity of each nonlinearity by itself.

\paragraph{Effect of number of images.} 
Fig.~\ref{fig:drums100} shows accuracy vs.\ number of epochs for the drums dataset when trained with all 100 images. WIRE results in highest accuracy within 2500 epochs and converges more rapidly than other approaches.
Fig.~\ref{fig:drums_nimg} shows accuracy as a function of number of training images.
%
%WIRE achieves highest PSNR across all training set sizes, with SIREN coming a close second by 0.1dB.
WIRE achieves 0.1dB higher than the next competitor SIREN for 25, 50, and 75 images, and 0.4dB higher when trained with 100 images.

Fig.~\ref{fig:drums} visualizes one of the reconstructed views for the drums. We varied the number of images from 25 to 100 and then rendered the image from a novel view.
Visually, WIRE generated the most pleasing results including sharp features of the cymbals and their stands, and the smooth membrane on the drum.
In contrast, Gaussian nonlinearity results in cloudy artifacts, while SIREN has high frequency artifacts, especially at lower numbers of images.
ReLU+positional encoding requires all 100 images and considerably more than 2500 epochs to reconstruct the components.
In all, WIRE is a a robust solution for training radiance fields, even with a small number of training samples.